\pdfoutput=1

\documentclass[11pt]{article}

\usepackage{EMNLP2022}

\usepackage{times}
\usepackage{latexsym}

\usepackage[T1]{fontenc}

\usepackage[utf8]{inputenc}

\usepackage{microtype}

\usepackage{inconsolata}

\usepackage{amsmath}
\usepackage{amssymb}
\usepackage{amsthm}
\usepackage{xcolor}
\usepackage{soul}
\usepackage{booktabs}
\usepackage{pifont}%
\usepackage{graphicx}
\usepackage{xspace}
\usepackage{enumitem}
\usepackage{caption}
\usepackage{subcaption}

\definecolor{mygreen}{RGB}{78, 159, 61}
\definecolor{myred}{RGB}{178, 6, 0}
\newcommand{\para}[1]{\noindent {\bf #1}}

\title{Active Example Selection for In-Context Learning}

\author{Yiming Zhang \and Shi Feng \and Chenhao Tan \\
        \tt{\{yimingz0, shif, chenhao\}@uchicago.edu} \\
        University of Chicago}

\DeclareMathOperator*{\argmax}{arg\,max}

\newcommand{\LM}[1]{\text{LM}_{#1}}
\newcommand{\pLM}[3]{\mathbf{P}_\text{LM}(#3 \vert #1 + #2)}

\newcommand{\secref}[1]{\S\ref{#1}}
\newcommand{\statstd}[2]{$#1_{#2}$}

\newcommand{\statespace}{\mathcal{S}}
\newcommand{\trset}{\mathbf{S}}

\newcommand{\state}{s}
\newcommand{\action}{a}

\newcommand{\actionspace}{\mathcal{A}}
\newcommand{\reward}{r}

\newcommand{\terminal}{\perp}

\newcommand{\qopt}{Q^\star}

\newcommand{\agnews}{\text{AGNews}\xspace}
\newcommand{\amazon}{\text{Amazon}\xspace}
\newcommand{\sst}{\text{SST-2}\xspace}
\newcommand{\trec}{\text{TREC}\xspace}

\newcommand{\sest}{\textsc{Seen examples, same task}\xspace}
\newcommand{\nest}{\textsc{New examples, same task}\xspace}
\newcommand{\nent}{\textsc{New examples, new task}\xspace}
\newcommand{\sametask}{\textsc{Same task}\xspace}
\newcommand{\newtask}{\textsc{New task}\xspace}

\newcommand{\gpt}{\textsc{GPT-2}\xspace}
\newcommand{\ada}{\textsc{Ada}\xspace}
\newcommand{\babbage}{\textsc{Babbage}\xspace}
\newcommand{\curie}{\textsc{Curie}\xspace}

\begin{document}
\maketitle

\begin{abstract}
    With a handful of demonstration examples, large-scale language models
    show strong capability to perform various tasks by {\em in-context}
    learning from these examples, without any fine-tuning.
    We demonstrate that in-context learning performance can be highly unstable
    across samples of examples, indicating the idiosyncrasies of how language models acquire information.
    We formulate example selection for in-context learning as a sequential decision
    problem, and propose a reinforcement learning algorithm for identifying
    generalizable policies to select demonstration examples.
    For GPT-2, 
    our learned policies demonstrate strong abilities of generalizing
    to unseen tasks in training, with a $5.8\%$ improvement on average.
    Examples selected from our learned policies can even achieve a small improvement on GPT-3 Ada. 
    However, the improvement diminishes on larger GPT-3 models, suggesting emerging capabilities of large language models.
\end{abstract}

\section{Introduction}

Large language models demonstrate the capability to learn from just a few
examples~\citep{radford2019language, brownLanguageModelsAre2020,
raeScalingLanguageModels2022, zhangOPTOpenPretrained2022}.
The possibility to train a model without any parameter update has inspired
excitement about the in-context learning paradigm.

Intuitively, high in-context learning performance should require
carefully chosen demonstration examples, but a recent line of work
suggests otherwise --- that demonstration examples are not as important as we
expected, and that few-shot performance can be largely attributed to the
model's zero-shot learning capacity~\citep{minRethinkingRoleDemonstrations2022}, across GPT-2 and GPT-3.
This insight is corroborated by a parallel line of work that brings
significant improvements to in-context learning performance without example
selection, for example, by re-ordering randomly selected examples and using calibration~\citep{luFantasticallyOrderedPrompts2022,zhaoCalibrateUseImproving2021, kojimaLargeLanguageModels2022}.
Another notable approach is to use best-of-$n$ sampling, which requires a labeled set
for validation~\citep{nakanoWebGPTBrowserassistedQuestionanswering2022}.

Our contribution in this paper is twofold.
First, we revisit the effect of example selection on in-context learning.
We show that even with reordering and calibration, we still observe a
large variance across sets of demonstration examples, especially for GPT-2, while calibration reduces the variance for GPT-3 models.
The high variance needs further
investigation, as we take it as evidence that large language models are still
not capable of efficiently and reliably acquire new information in-context.
Understanding what makes good demonstration examples sheds some light on the
 mechanisms that large language models use to process information.

Second, we seek to discover general trends in example selection for
in-context learning across different tasks.  Concretely, we use reinforcement
learning to optimize example selection as sequential decision making problem.
We argue that active example selection from unlabeled datasets is the most appropriate setting for in-context learning because fine-tuning with an existing labeled set leads to great performance with low variance.
For GPT-2, we validate our learned policy on a seen task with labeled dataset and observe a 12.1\% improvement over a max-entropy active learning baseline.
Moreover, our learned policy is able to generalize to new tasks with 5.8\% improvement, suggesting that the policy is able to capture systematic biases in how GPT-2 acquires information.
Examples selected from our learned policies can even achieve a small improvement on GPT-3 Ada. 
However, the improvement diminishes on larger GPT-3 models.
We provide further analyses to understand the properties of useful examples.

Overall, our work explores how large language models process information through the perspective of example selection and formulate active example selection as a sequential decision making problem.
We investigate divergent behaviors between GPT-2 and GPT-3, which echoes the emerging abilities of large language models, and suggest that researchers in the NLP community should collectively build knowledge and research practice in the era of large language models.\footnote{Our code is available at \url{https://github.com/ChicagoHAI/active-example-selection}.}

\section{The Effect of Example Selection}
\label{sec:am}

In this section, we demonstrate the instability of in-context learning
performance due to the selection of demonstration examples.
We further show that existing methods (e.g., calibration,
reordering) are insufficient for addressing this stability for GPT-2.
In comparison, the variance of GPT-3 models can be mitigated with
calibration.

\subsection{In-context Text Classification with Demonstration Examples}
\label{sec:in_context_example_selection}

We start by formally defining in-context learning.
We focus on in-context learning for text classification with a left-to-right
language model.
All supervision is given through a ``prompt'' which we denote as $\state$.
The prompt typically contains natural language instructions and a few demonstration examples.
To make a prediction for a test example $x$, we concatenate the prompt and
the test example as prefix, and use the language model to predict the next
token: $\argmax_{y} \mathbf{P}_\text{LM}(y|\state+x)$, where $+$ denotes concatenation.
Typically, instead of taking the $\argmax$ from the whole vocabulary, we
restrict the model's output to a set of special tokens which corresponds to the
set of labels, e.g., with the word ``positive'' corresponding to the positive
class in binary sentiment classification.
In our formulation, we omit a separate variable for the special tokens, 
and use $\mathcal{Y}$ to refer to
both the label set and the set of proxy tokens for simplicity.

To summarize, a prompt in this paper is a sequence of $k$ {\bf labeled}
examples concatenated together: $\state=(x_1, y_1), (x_2, y_2), \dots, (x_k, y_k)$.
And the prediction for a test input $x$ is the label with the highest likelihood
of being by the language model: $\argmax_{y\in\mathcal{Y}} \pLM{\state}{x}{y}$.\footnote{If a label is represented as multiple tokens in the LM,
e.g., \texttt{negation}=\texttt{neg}+\texttt{ation}, we simply use the
first unambiguous token, e.g., \texttt{neg} for \texttt{negation} and
\texttt{ent} for \texttt{entailment}.}

\paragraph{Experiment setup.} Following \citet{zhaoCalibrateUseImproving2021},
we conduct our experiments on
\agnews~\citep{zhangCharacterlevelConvolutionalNetworks2015},
\sst~\citep{socherRecursiveDeepModels2013} and
\trec~\citep{voorheesBuildingQuestionAnswering2000}.
We additionally include
\amazon~\citep{zhangCharacterlevelConvolutionalNetworks2015} since it contains
longer texts than the remaining datasets.
Table~\ref{tab:stats} give basic information of the tasks.

Using {GPT-2 345M} (\gpt), GPT-3 Ada (\ada) and
GPT-3 Babbage (\babbage) as the in-context learning models, we report $4$-shot
example selection performance across all experiments.

\begin{table}[t]
  \small
  \centering
  \begin{tabular}{@{}llrr@{}}
  \toprule
  Dataset & Domain                & \#classes & avg. length    \\ \midrule
  \agnews & Topic cls.            & 4 & 37.8              \\
  \amazon & Sentiment cls.        & 2 & 78.5              \\
  \sst    & Sentiment cls.        & 2 & 19.3              \\
  \trec   & Question type cls.    & 6 & 10.2              \\  \bottomrule
  \end{tabular}
  \caption{Dataset information.}
  \label{tab:stats}
\end{table}

\begin{figure}[t]
  \centering
  \includegraphics[width=0.4\textwidth]{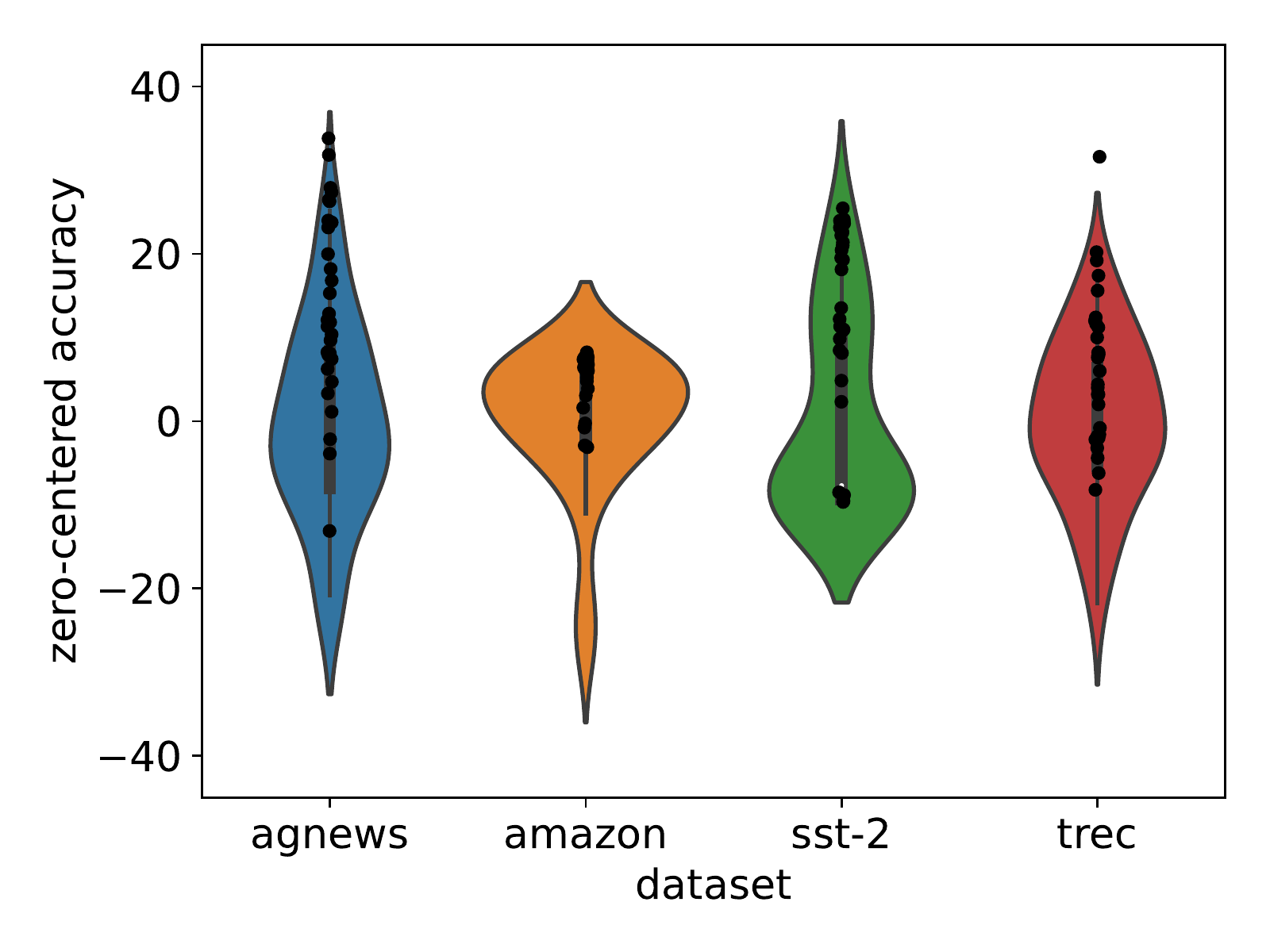}
  \caption{Zero-centered in-context learning accuracy of \gpt on 30 random sets of 4 demonstration
  examples. Each dot indicates performance of the best permutation for one set
  of demonstration examples.
  $y$-axis represents the accuracy difference with the mean accuracy of random demonstration examples.}
  \label{fig:random-sampling}
\end{figure}

\subsection{Sensitivity to Example Selection}

We first highlight the sensitivity of GPT-2 due to example selection.
In Figure~\ref{fig:random-sampling}, we plot the in-context learning performance
of 30 random sequences of demonstration examples with length 4.
Across all 4 tasks, the maximum and minimum performance due to random
sampling differs by $>30\%$.
Additionally, for 3 out of the 4 tasks (\agnews, \sst and \trec), performance of
the worst set of demonstration examples lead to in-context learning performance
below random guessing (e.g., it is $10.0\%$ on \trec, below $16.7\%$ accuracy of
guessing randomly among 6 labels in \trec).

\paragraph{Reordering sequence alone cannot address the instability.} \citet{luFantasticallyOrderedPrompts2022} identifies the ordering of demonstration
examples as the cause for variance, and proposed heuristics to reorder
demonstration examples.
For such an approach to be effective, the underlying assumption is that there
exists good orderings for most sets of demonstration examples.

In Figure~\ref{fig:random-sampling}, we additionally report the highest possible
performance among $4!=24$ permutations for each of the 30 sets 
using a validation set of 100 examples.
The reordering performance reported here is highly optimistic
for a true few-shot setting \citep{perezTrueFewShotLearning2021} since a
validation set cannot be assumed available.
As expected, taking the best permutation on a validation set improves test
performance: we observe an average of 8.1\% increase on average over random
demonstration examples.

However, these best orderings of examples still lead to a wide range of
possible performance. 
On \agnews, we observe a maximum accuracy of $79.6\%$ and a minimum accuracy of
$32.7\%$ after considering the best possible orderings.
On \trec, the best ordering for 9 out of 30 sets of examples lead to performance
below random examples.
These observations suggest that there are simply no good orderings for
considerable proportions of demonstration sets, motivating the need for
selecting examples beyond merely reordering.

\begin{figure}[t]
  \centering
  \includegraphics[width=0.4\textwidth]{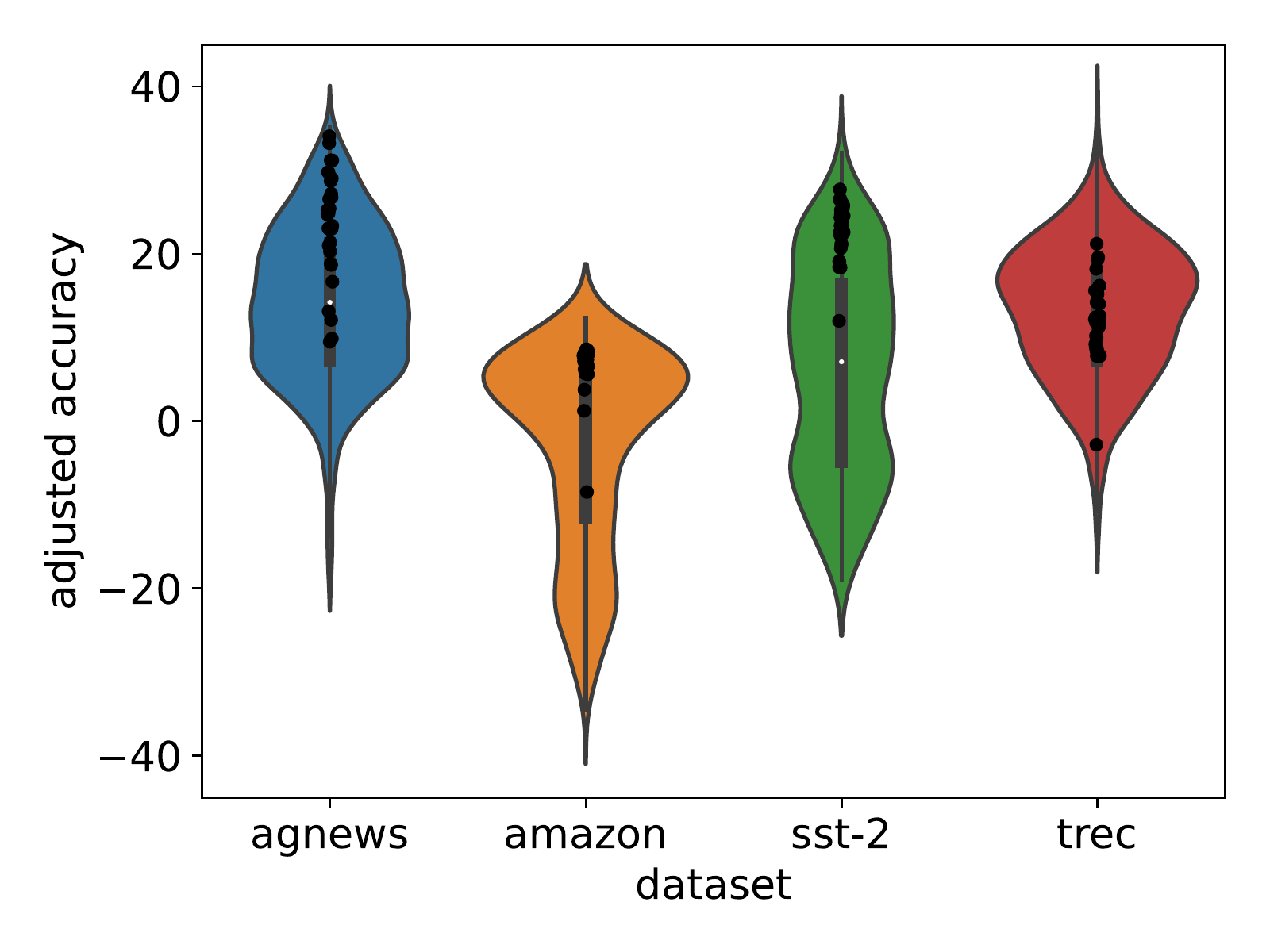}
  \caption{In-context learning accuracy of 30 random sets of 4 demonstration
  examples \textbf{with calibration}.
  Each dot indicates performance of the best permutation for one set of
  demonstration examples.
  Accuracy over random examples (\textbf{no calibration}) is plotted.}
  \label{fig:random-sampling-calibrate}
\end{figure}

\paragraph{Calibration does not decrease variance for GPT-2, either.}
\citet{zhaoCalibrateUseImproving2021} finds that language models 
are 
poorly
calibrated when used directly as in-context classifiers, and argues that
calibration is the key missing piece to improve and stablize in-context learning
performance.
It proposes using dummy examples (e.g., ``N/A'') as anchors for calibrating
the language model since a calibrated language model should make neutral predictions for these content-free examples.

Figure~\ref{fig:random-sampling-calibrate} demonstrates the effectiveness of
calibration in improving few-shot performance.
With calibration, we observe an increase in average performance of varying
magnitude on 3 out of the 4 tasks (\agnews, \sst and \trec), but a marginal
decrease of performance on \amazon.
For example, on \agnews where calibration improves performance the most, we
observe a maximum accuracy of $79.5\%$ and a minimum accuracy of $26.1\%$,
resulting in a gap of over $53.4\%$.

Interestingly, we observe varying behavior when combining calibration with
demonstration reordering.
On the binary tasks (\amazon and \sst), we observe prompt reordering to be
quite effective, consistently leading to performance above random examples.
On the other hand, for \agnews (4 labels) and \trec (6 labels), we observe
much greater variance.

In summary, with GPT-2, existing methods do not provide satisfactory solutions to the sensitivity of in-context learning to demonstration examples.
Reordering demonstration requires a well-behaving demonstration set, which is
often not the case, and does not reduce variance.
Calibration, though improves performance, does not reduce variance, and its
effectiveness deteriorates with a large label set.
These findings motivate the need for identifying high quality demonstration
examples for consistent and performant in-context learning.

\begin{table}[t]
    \centering
    \small
    \begin{tabular}{@{}lllll@{}}
        \toprule
        Model               & \agnews               & \amazon               & \sst                  & \trec                 \\ \midrule
        \gpt                & \statstd{44.5}{9.3}   & \statstd{87.5}{3.7}   & \statstd{61.7}{14.4}  & \statstd{29.4}{12.8}  \\
        \gpt ({\bf C})      & \statstd{55.2}{12.0}  & \statstd{76.3}{14.0}  & \statstd{66.2}{14.7}  & \statstd{40.8}{5.4}   \\ \midrule
        \ada                & \statstd{62.9}{17.5}  & \statstd{87.0}{6.1}   & \statstd{65.0}{10.2}  & \statstd{21.2}{6.6}   \\
        \ada ({\bf C})      & \statstd{64.0}{4.0}   & \statstd{90.0}{1.2}   & \statstd{73.8}{9.7}   & \statstd{22.1}{5.3}   \\ \midrule
        \babbage            & \statstd{68.0}{14.0}  & \statstd{93.4}{0.8}   & \statstd{92.2}{2.7}   & \statstd{27.4}{5.8}   \\
        \babbage ({\bf C})  & \statstd{78.1}{6.1}  & \statstd{92.7}{1.6}  & \statstd{90.8}{1.1}     & \statstd{36.0}{4.0}   \\
        \bottomrule
        \end{tabular}
    \caption{Performance of \gpt, \ada and \babbage
    across 5 random sets of 4-shot demonstration examples. {\bf C} indicates
    calibration. Standard deviation is reported as subscripts.}
    \label{table:gpt3}
\end{table}

\paragraph{Variance persists to some degree with GPT-3.}
In Table~\ref{table:gpt3}, we report the performance of \gpt,
\ada and \babbage on 5 random sets of demonstration examples.\footnote{We do not use the same sample size or examine the effect of re-ordering for cost considerations.}
GPT-3 models are not immune to instability due to resampling demonstration
examples.
On multi-labeled tasks including \agnews and \trec, we observe both \ada and
\babbage demonstrate significant variance, and on binary tasks such as \amazon
and \sst, much smaller variance is observed.
This difference is potentially due to the difficulty of the task and the multi-class nature of \agnews and \trec.
We will address the latter in \secref{sec:analysis}.
Another interesting observation is that variance diminishes with calibration.
However, one may argue that calibration no longer reflects the model's innate ability to acquire information.

Overall, the differences in model behavior between GPT-2 and GPT-3 add evidence to the emergent ability of large language models \citep{wei2022emergent,bowman-2022-dangers}.
We hypothesize that the variance will be even smaller with GPT-3 Davinci.

\section{Active Example Selection by RL}
\label{sec:framework}

Given a set of {\em unlabeled} examples, can we choose the right ones to be
 annotated as demonstration examples?
In this section, we formulate the problem of active example selection for
 in-context learning.
Following the definition of in-context learning in
 \secref{sec:in_context_example_selection}, constructing a prompt for in-context
 learning boils down to choosing a sequence of demonstration examples.

We emphasize that by selecting from {\em unlabeled} examples, our setup is
 analogous to active learning, where we select examples to label.
We think that this is the most appropriate setting for in-context learning
 because fine-tuning can lead to great performance with low variance if we
 already have a moderately-sized labeled set (e.g., 100 instances).

As in-context learning uses a small number of examples, we formulate active
 example selection as a sequential decision making problem, where prompt is
 constructed by selecting and annotating one demonstration example at a time.
We use a Markov Decision Process (MDP) to formalize the problem, discuss our
 design of the reward function, and introduce our solution to example selection
 using reinforcement learning (RL).

\subsection{Active Example Selection as a MDP}

Given a set of unlabeled examples, we want to maximize the expected accuracy on
 unseen test examples by getting up to $k$ annotations.
The space of possible prompts grows exponentially with the number of unlabeled
 example and is intractable to enumerate, so we treat it as a sequential
 decision making problem: given the pool of unlabeled examples
 $\trset_\mathcal{X}=\{x_i\}$, choose one example $x_i$, obtain its groundtruth
 label $y_i$, append the pair $(x_i, y_i)$ to our prompt, and repeat this
 process until either the budget $k$ is exhausted or the policy takes a special
 action $\terminal$ indicating early termination.

\para{Action space and state space.}
The action space of the MDP is the set of unlabeled examples plus the special
 end-of-prompt action: $\actionspace = \trset_\mathcal{X} \cup \{\terminal\}$.
After choosing an action $x_i$ we observe its label $y_i$, and the state is
 defined by the prefix of the prompt $\state=(x_1, y_1), (x_2, y_2), \dots,
	 (x_i, y_i)$.

\para{Reward.}
The reward $r$ can be defined based on an arbitrary scoring function $f$ of the
 language model $\LM{}$ when conditioned on the prompt $s$, denoted $r =
	 f(\LM{\state})$.
In practice, we use the accuracy on a labeled validation set as reward.

It follows that we need to have access to a validation set during training,
 which we refer to as {\em reward set}.
Similarly, we also have a labeled set from which our policy learns to select
 examples.
We refer to this labeled set as {\em training set}.
Ideally, our learned policies identify generalizable qualities of demonstration
 examples and can select useful unlabeled examples in a task where the policy
 has not observed any labeled examples.
We will explore different setups to evaluate our learned policies.

It is useful to emphasize how active example selection  deviates from the
 standard reinforcement learning setting.
First, the action space is the examples to be selected, which can be variable
 in size.
Furthermore, the actions during test time can be actions that the policy has
 never observed during training.
Similarly, the classification task can differ from training, analogous to a new
 environment.
Such generalizations are not typically assumed in reinforcement learning, due
 to the challenging nature of the
 problem~\citep{kirkSurveyGeneralisationDeep2022}.

\subsection{Active Example Selection by Q-learning}

Framing active example selection as a sequential problem allows us to use
 off-the-shelf RL algorithms to train a policy.
We opt to use Q-learning~\citep{mnihPlayingAtariDeep2013} for its simplicity
 and effectiveness.

The objective of Q-learning is to approximate the optimal state-value function
 $Q^\star(s, \action)$, i.e., the maximum (discounted) future reward after
 taking action $a$ in state $s$.
The Bellman equation \cite{bellmanDynamicProgramming1957} allows a recursive
 formulation of the optimal state-value function $\qopt$ as $$ \qopt(s, \action)
 = \mathbb{E}_{s \sim \statespace} \left[ \reward(s, \action) + \gamma
	 \max_{\action'} \qopt(s', \action') \right] \text{.
}
$$

We collect off-policy training data in our implementation and thus use offline
 Q-learning to leverage off-policy
 data~\citep{prudencioSurveyOfflineReinforcement2022}.
Specifically, We use conservative Q-learning (CQL)
 ~\citep{kumarConservativeQLearningOffline2020}, which uses regularization to
 prevent the overestimation of Q-values for unobserved actions in training data,
 contributing to a robust policy when evaluated in an unfamiliar environment.
More details about CQL can be found in the Appendix~\ref{appendix:cql}.

\para{Generation of off-policy data.}
Offline learning requiers off-policy training data.
We run a random policy for a fixed number (2,000) of episodes to create the
 off-policy data.
For every episode, we randomly sample 4 demonstration examples, and compute
 features and intermediate rewards.
Then, we store the trajectory as training data.

\para{Feature-based representation of actions.}
In our framework, a state $\state$ is a sequence of examples, and we simply use
 the number of already selected examples $\lvert \state \rvert$ as the feature
 representation.
To enable our method to be deployed in an active example selection process, we
 assume no access to labels prior to selecting an example.
That is, when representing a example to be selected $\action = (x, y)$, we omit
 the label $y$ and simply use predicted label probabilities conditioned on the
 current examples $\pLM{\, \state}{x}{\, \cdot\, }$.
We additionally include entropy of the prediction.%
\footnote{Other features can
	be used, such as embeddings of the language model.
	We use minimal features so that policies could be evaluated across models
	 (GPT-2 and GPT-3).
}

\para{Reward shaping.}
The previously defined reward function only rewards a completed prompt, while
 intermediate states receive zero reward.%
Sparse reward schemes are known to make learning
 difficult~\citep{pathakCuriosityDrivenExplorationSelfSupervised2017}.
Therefore, we propose an alternative reward function based on the marginal
 utility of actions \citep{von1893natural}.
At time step $t$ we define $\reward : \statespace \times \actionspace \to
	 \mathbb{R}$ as \[ r(\state, \action) = f(\LM{\state + \action}) -
 f(\LM{\state}) \text{.
}
\]

Intuitively, $r$ measures the ``additional gain'' on objective $f$ by acquiring
 the label of example $a$.
Notice that $f(\LM{\varnothing})$ can be conveniently interpreted as the
 zero-shot performance of the language model.
Maximizing this marginal utility reward function is indeed equivalent to
 optimizing the true objective $f$: observe that the summation of rewards along
 a trajectory is a telescoping series, leaving only the final term
 $f(\LM{s_\terminal})$ minus a constant term that does not affect the learned
 policy.%
\footnote{Requires
	the discount factor $\gamma = 1$, which we use in across all
	experiments.}
It turns out that $r$ is a {\bf shaped
		 reward}~\citep{ngPolicyInvarianceReward1999}, a family of transformed reward
 functions that preserves the invariance of optimal policies.

\para{Target network with replay buffer.}
Our algorithm uses separate policy and target
 networks~\citep{hasseltDoubleQlearning2010} with a replay
 buffer~\citep{linSelfImprovingReactiveAgents1992}.
Both are standard extensions to vanilla DQN
 \citep{arulkumaranBriefSurveyDeep2017}, and are demonstrated to improves
 performance while alleviating certain optimization
 issues~\citep{hesselRainbowCombiningImprovements2017}.
After concatenating state and action representations, we use a 3-layer MLP as
 the Q-network: $\hat{Q}(\state, \action) = \text{MLP}([ \state \mathbin\Vert
		 \action ])$.
We report hyperparameters details in Appendix \ref{appendix:hyperparameters}.

\section{Results}

In this section, we investigate the performance of our learned policies for
 \gpt.
Due to the significant costs of generating episodes, we only apply the policies
 learned from GPT-2 and examine direct transfer results on GPT-3.
Baselines, oracles and our method have access to the same underpinning calibrated
 GPT-2 model.

\begin{table*}[t]
    \small
    \centering
    \begin{tabular}{@{}lccccc@{}}
    \toprule
    Method                                  & Average   & \agnews         &\amazon        & \sst          & \trec         \\ \midrule
    random                                  & $59.6$    & $55.2_{10.5}$   & $76.3_{12.3}$ & $66.2_{12.9}$ & $40.8_{4.7}$  \\
    max-entropy                             & $59.3$    & $58.8_{11.3}$   & $74.8_{5.1}$  & $65.7_{10.7}$ & $37.8_{6.7}$  \\ 
    reordering                              & $63.5$    & $63.3_{6.8}$    & $89.8_{3.8}$  & $67.9_{11.1}$ & $33.0_{4.2}$  \\ \midrule
    best-of-10                              & $72.5$    & $72.1_{1.9}$    & $91.1_{0.6}$  & $81.1_{4.4}$  & $45.6_{3.5}$  \\
    greedy-oracle                           & $78.0$    & $80.6_{1.7}$    & $91.8_{1.1}$  & $81.7_{3.9}$  & $58.0_{7.5}$  \\ \midrule
    our method ({\bf seen examples})        & $71.4$    & $70.8_{7.8}$    & $90.4_{1.9}$  & $81.0_{3.5}$  & $43.3_{2.0}$  \\
    our method ({\bf 100 new examples})     & $71.6$    & $71.3_{7.4}$    & $89.2_{3.9}$  & $81.8_{2.6}$  & $44.0_{4.6}$  \\
    our method ({\bf 1000 new examples})    & $69.0$    & $65.5_{7.4}$    & $88.5_{4.2}$  & $76.7_{7.5}$  & $45.4_{5.0}$  \\ \bottomrule
    \end{tabular}
    \caption{{\bf \sametask} accuracy on \agnews, \amazon, \sst and \trec,
    across 5 random seeds.
    $95\%$ confidence intervals are reported as subscripts.}
    \label{table:main}
\end{table*}

\subsection{Setup}

Following our framework in \secref{sec:framework}, during training, we use a
 \textbf{training set} from which the trained policy picks 4 examples for
 demonstration, as well as a \textbf{reward set}, which is a validation set
 where we compute rewards for the learning agent.
Each set has 100 examples and our training scheme uses a total of 200 examples.

Depending on the availability of a reward set, we consider three evaluation
 settings:  \begin{itemize}[leftmargin=*,itemsep=0pt] \item {\bf \sest.
}
In this setting, we use the learned policy to pick demonstration examples from
 the \textbf{training set}.
We expect our method to be competitive with oracle methods that select examples
 based on rewards.

\item {\bf \nest.}
We consider a more challenging setting where the learned policy picks from an
 \textbf{unlabeled set} of 100 or 1000 previously unseen examples.
The learned policy still benefits from access to the reward set during
 training as the classification task is the same, but it cannot perform well
 simply by memorizing good sequences.

\item {\bf \nent.}
Finally, we ask the learned policy to pick examples on a new task that it has
 never seen.
Specifically, we adopt a multi-task learning approach, allowing the policy to
 simultaneously learn from all but one tasks.
Then, we evaluate the held-out task (e.g., train on \agnews, \sst, \trec and
 test on \amazon).
The learned policies use 600 examples from training (3 $\times$ 100 each for
 the \textbf{training set} and \textbf{reward set}).
During evaluation, the policy picks examples from an \textbf{unlabeled set} of
 examples in the held-out task, and we experiment with either 100 or 1000 unlabeled
 examples.
\end{itemize}

\sest and \nest serve as sanity check of our learned policies, while \nent is
the most appropriate setting for evaluating in-context learning.

\paragraph{Baselines and oracles.}
We consider three baseline methods for example selection.
The \textbf{random} strategy simply picks demonstration examples randomly.
Our second baseline (\textbf{max-entropy}) is a standard approach in active
 learning~\citep{settlesActiveLearningLiterature2009,
	 daganCommitteebasedSamplingTraining1995} which greedily picks the example
 maximizing classification entropy.
We additionally consider a strong example reordering heuristic by
 \citet{luFantasticallyOrderedPrompts2022} , dubbed {\bf reordering};\footnote{\citet{luFantasticallyOrderedPrompts2022} experiment with two metrics
 for selecting the best ordering.
In the {\bf reordering} baseline, we use the ``Global Entropy'' metric since
 it performs better on average in the original paper.
}
{\bf reordering} first uses the language model to generate a set of fake examples that
resemble demonstration, and then chooses an ordering that
maximizes classification entropy on these fake examples.
Intuitively, {\bf max-entropy} and {\bf reordering} both encourages class
 balance during prediction.
All three baselines can be used in active example selection, namely, example
 selection that does not have label access to examples before they are selected.

We further consider two oracle methods that require a labeled candidate set and
 a reward set.
The \textbf{best-of-10} strategy randomly samples 10 times and keeps the sample
 that maximizes performance on the reward set as the final demonstration
 sequence.
In addition,  we use a greedy strategy to iteratively choose the example that
 results in the highest performance on the reward set, and we refer to this
 strategy as \textbf{greedy-oracle}.
The oracles do not work for active example selection and cannot be used in
 \newtask as the assumption is that we do not have any labeled examples, so we
 do not compare our learned policies with oracles in \newtask.

We use baselines and our methods to select 4 demonstration examples for every
 task, and we average model performances across 5 random runs.

\subsection{Main results}
\label{sec:main-results}
\begin{table*}[t]
    \centering
    \begin{tabular}{@{}lccccc@{}}
    \toprule
    Method                              & Average   & \agnews         &\amazon        & \sst          & \trec         \\ \midrule
    random                              & $59.6$    & $55.2_{10.5}$   & $76.3_{12.3}$ & $66.2_{12.9}$ & $40.8_{4.7}$  \\
    max-entropy                         & $59.3$    & $58.8_{11.3}$   & $74.8_{5.1}$  & $65.7_{10.7}$ & $37.8_{6.7}$  \\
    reordering                          & $63.5$    & $63.3_{6.8}$    & $89.8_{3.8}$  & $67.9_{11.1}$ & $33.0_{4.2}$  \\ \midrule
    our method ({\bf 100 examples})     & $63.8$    & $63.4_{10.4}$   & $86.8_{6.7}$  & $65.9_{13.4}$ & $38.9_{5.1}$  \\
    our method ({\bf 1000 examples})    & $65.4$    & $66.7_{5.7}$    & $89.9_{1.6}$  & $61.9_{7.7}$  & $43.3_{4.4}$  \\ \bottomrule
    \end{tabular}
    \caption{\textbf{New-task} accuracy on \agnews, \amazon, \sst and \sst,
    across 5 random seeds. $95\%$ confidence intervals are reported as
    subscripts.}
    \label{table:transfer}
\end{table*}

We analyze the effectiveness of applying our method in both  \sametask and
 \newtask.%

\para{\sametask.}
Our method evaluated by picking from {\bf seen examples} demonstrates strong
 performance.
Across all 4 tasks, our method outperforms random, max-entropy and reordering
 baselines by an average of $11.8\%$, $12.1\%$ and $7.9\%$, respectively, as well as $>10\%$
 improvements on 2 tasks.

Beyond performance gains, it is clear that our method helps reduce variance.
We present $95\%$ confidence intervals as a proxy for variance.
Across all 4 tasks, we observe consistent decrease in variance compared to the
 baselines.

Picking from both 100 and 1000 {\bf new examples} largely retains the
 performance gains and variance reductions.
Interestingly, we notice a higher overall performance of picking from 100 over
 1000 new examples.
This can be attributed to the large variance (see
 Appendix~\ref{sec:unlabeled-size} for more results).

Comparing with oracle methods, our methods perform relatively closely to {\bf
		 best-of-10}, while {\bf greedy-oracle} significantly outperforms the other
 methods.
Since we want the policies to learn generalizable example selection strategies,
 we intentionally use simple features, which may explain why our method, even
 when picking from seen examples, does not outperform oracles.
Thanks to the high variance of random sampling, {\bf best-of-10} is a
 very performant strategy despite its simplicity, and a reasonable choice
 if validation is possible.
At the cost of an exponential runtime,
{\bf greedy-oracle} shows the strong
 in-context learning performance attainable with just example selection,
 motivating the framing of in-context learning optimization as a pure example
 selection problem.
In fact, the average performance from {\bf greedy-oracle} with GPT-2 (345M)
is better than that of GPT-3 Curie, a 20x larger model (see
 Appendix~\ref{sec:gpt3-transfer}).\footnote{The sizes of GPT-3 models hosted by OpenAI are not
 publicly known, and we use estimations at \url{https://blog.eleuther.ai/gpt3-model-sizes}.}

\para{\newtask.}
We further evaluate our methods under the new task setting, where we train the
 example selection policy on 3 tasks, and evaluate on a previously unseen task.
On average, we observe a smaller, but still significant improvements over both
 random and max-entropy baselines, suggesting the existence of learnable
 insights about good demonstration examples that generalize across tasks.
On the other hand, we observe limited gains over reordering, signifying the
 challenge of finding good examples in an unknown task.

Interestingly, when picking from 1000 examples, we observe a much greater
 effect of variance reduction compared to baselines.
In comparison, the variance reduction effect is minimal when picking from 100
 examples and the performance gain is slightly smaller likely due to randomness.

We continue this discussion on the effect of size of selection set on transfer
 performance in Appendix~\ref{sec:unlabeled-size}.

\para{GPT-3 transfer.}
Training example selection policies directly on GPT-3 models is not viable
 since it requires sample a significant number of trajectories while computing
 rewards.
Therefore, we instead evaluate if policies and examples trained on GPT-2
 generalize to GPT-3.
Overall, we find mixed transfer results.
On the smaller GPT-3 \ada model, we observe small gains ($\sim 1\%$) by
 transferring both policies and examples, which is impressive consider the
 architectural differences between GPT-2 and GPT-3.
However, we observe mixed results in transfer to \babbage and \curie.
We report further details in Appendix~\ref{sec:gpt3-transfer}.

\begin{figure*}[t]
    \centering
    \begin{subfigure}[b]{0.38\textwidth}
        \centering
        \includegraphics[width=\textwidth]{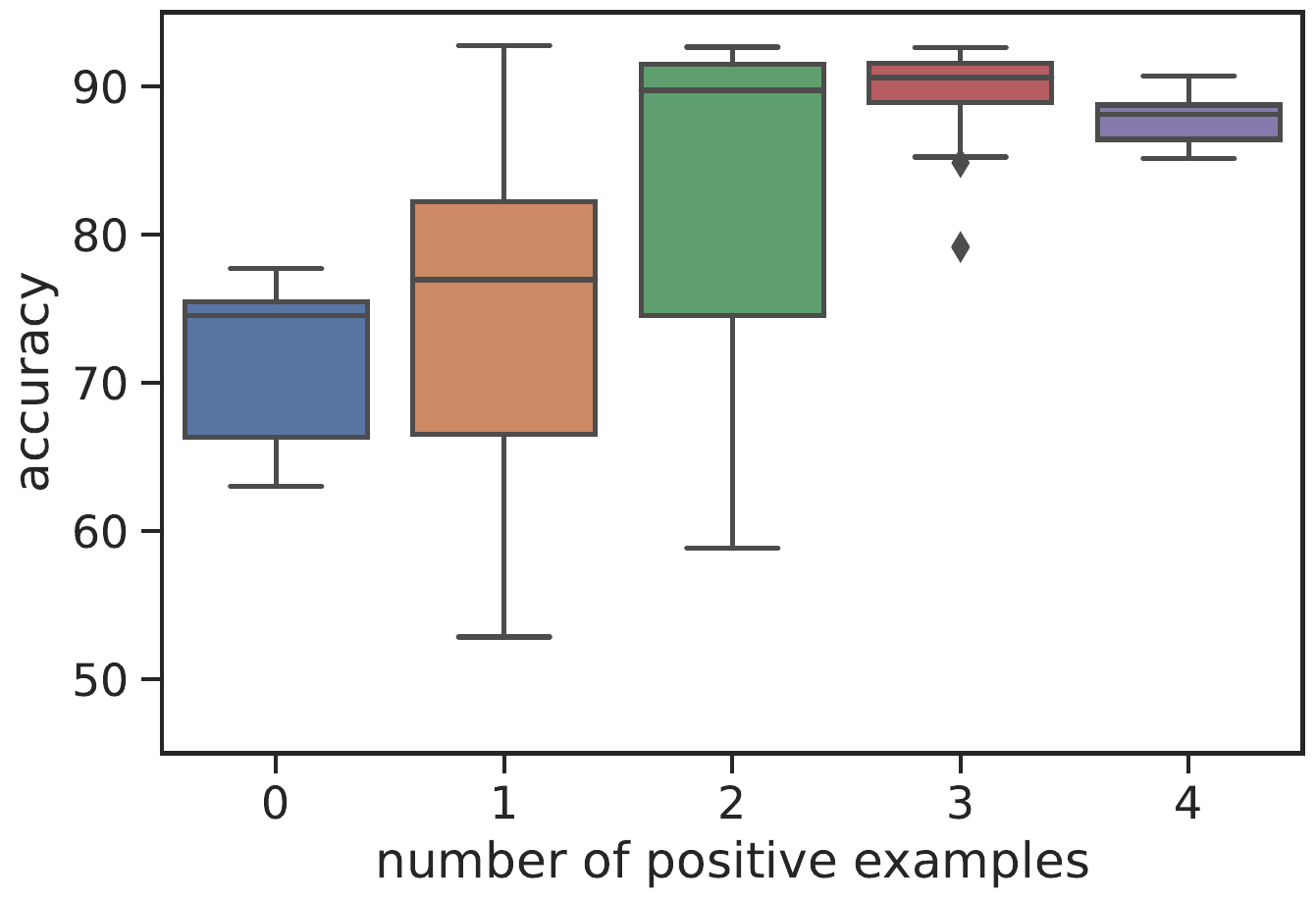}
        \caption{\amazon}
    \end{subfigure}
    \begin{subfigure}[b]{0.38\textwidth}
        \centering
        \includegraphics[width=\textwidth]{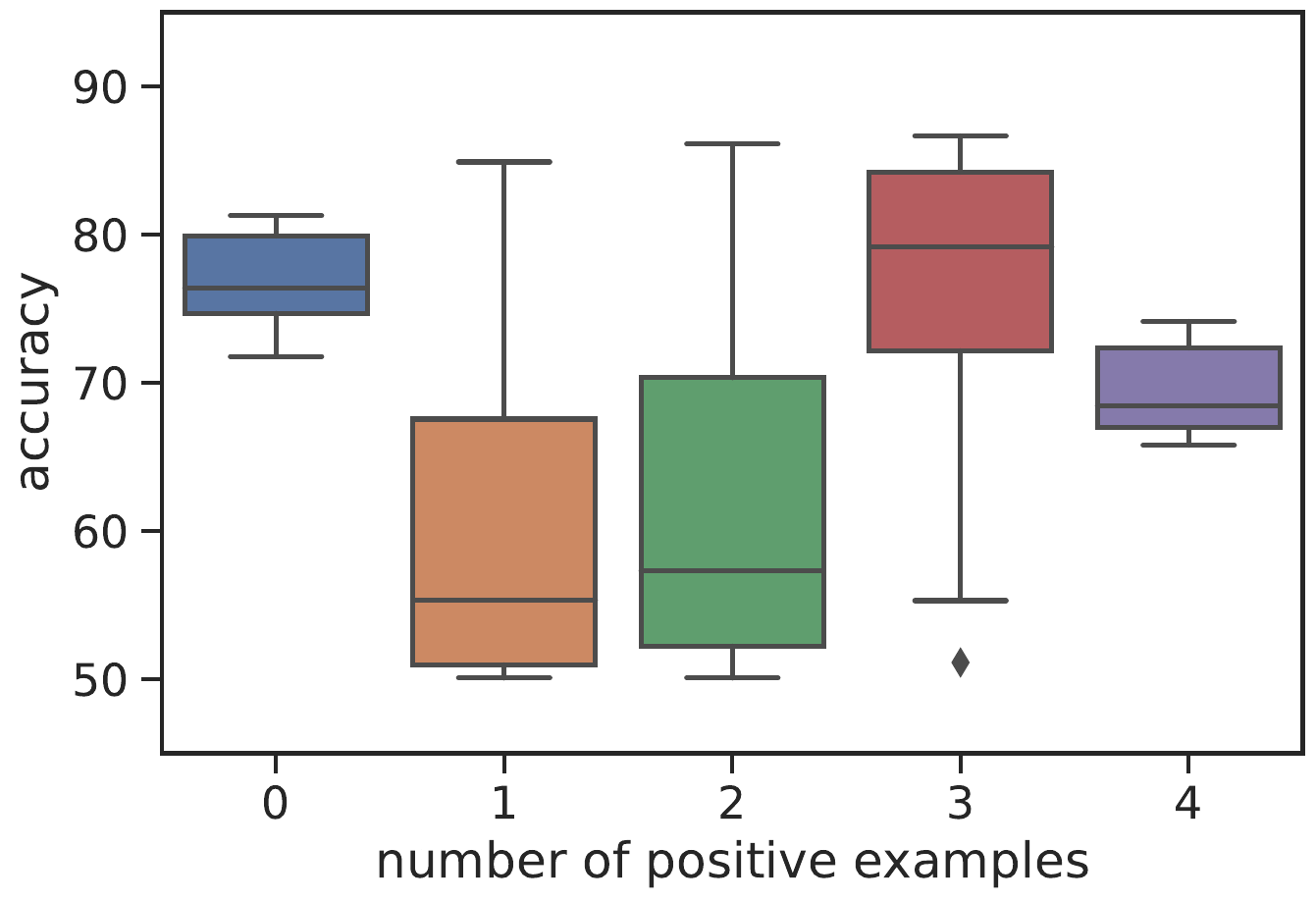}
        \caption{\sst}
    \end{subfigure}
    \caption{Accuracies of \amazon and \sst with varying {\bf label balance}
    (number of positive examples in demonstration), across 100 total random
    samples of 4 demonstration examples.}
    \label{fig:balance}
\end{figure*}

\begin{figure*}[t]
    \centering
    \begin{subfigure}[b]{0.38\textwidth}
        \centering
        \includegraphics[width=\textwidth]{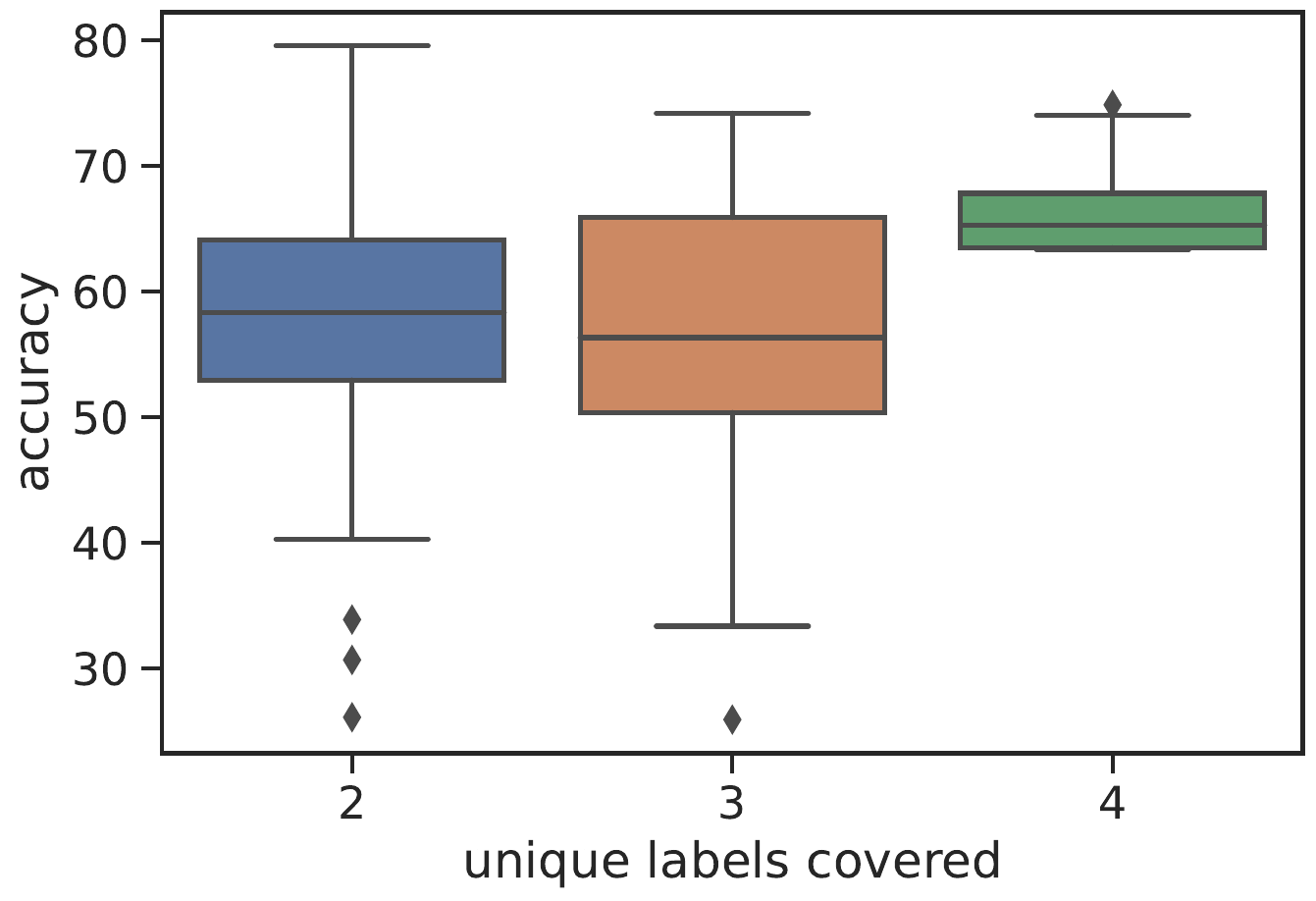}
        \caption{\agnews}
    \end{subfigure}
    \begin{subfigure}[b]{0.38\textwidth}
        \centering
        \includegraphics[width=\textwidth]{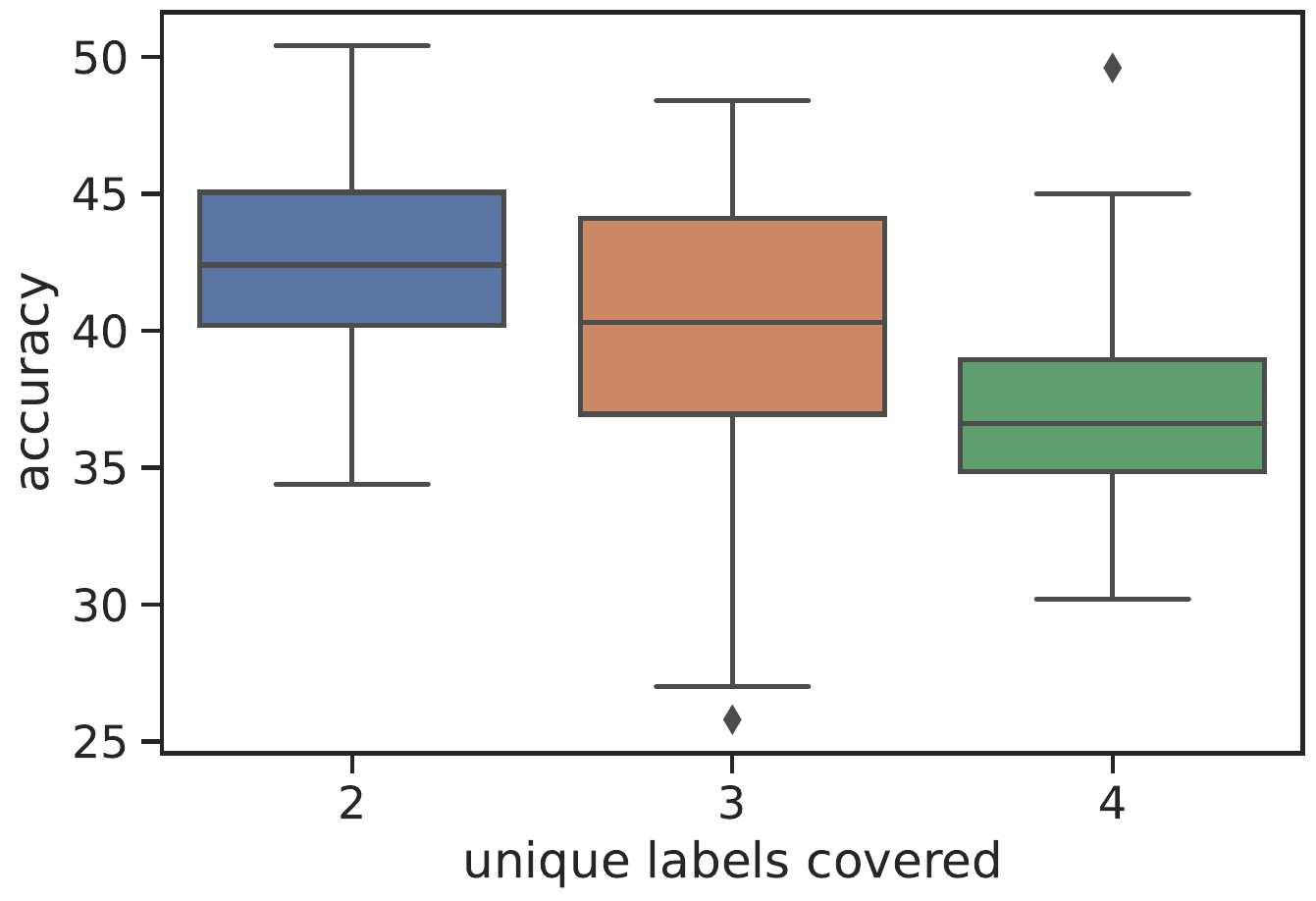}
        \caption{\trec}
    \end{subfigure}
    \caption{Accuracies of \agnews and \trec with varying {\bf label coverage}
    (number of unique labels covered in demonstration), across 100 total random
    samples of 4 demonstration examples. Demonstration set that only covers 1
    label is very unlikely and does not appear in our experiments.}
    \label{fig:coverage}
\end{figure*}

\subsection{What Makes Good Examples?}
\label{sec:analysis}

To understand what makes good examples, we explore properties of the learned policy and design additional experiments based on our qualitative examination of the selected examples.
In the interest of space, we focus on label balance and coverage, and present other results based on linear policies (\ref{appendix:linear}) and length (\ref{appendix:length}) in the Appendix.

On \amazon and \sst, both binary sentiment classification tasks, we focus on
label balance, measured by the number of {\bf positive} labels in the
demonstration set.
For \agnews (4 labels) and \trec (6 labels), we instead focus on the distinct
number of labels covered in demonstration.
We present the results in Figure~\ref{fig:balance} and Figure~\ref{fig:coverage}.

Perhaps surprisingly, a well-balanced demonstration set does not consistently
lead to greater performance or less variance.
In \amazon, we notice that having all 4 examples being positive actually leads
to good in-context learning performance, with an average accuracy of $87.8\%$
and $4.5\%$ greater than that of a perfectly balanced demonstration set
($83.3\%$).
A similar trend is demonstrated in \sst, where having all positive or all
negative labels leads to much smaller variance compared to more balanced sets,
while outperforming perfectly balanced sets on average.

In \trec, we again observe that the model does not need to observe the entire
label space to perform well.
The greatest performance occurs when exactly two labels are covered by
demonstration, and the performance deteriorates as label coverage increases.
\agnews demonstrates a somewhat expected pattern. When 4 label are covered,
we observe the best performance along with a small variance.
That said, covering three labels does not improve over covering two labels.

Overall, our analysis highlights the idiosyncrasies of how GPT-2 acquires information in in-context learning.
The sequences that lead to strong performance may not align with human intuitions.

\section{Related Work}

Our paper builds on top of prior work
that uses RL to solve the active learning problem~\citep{fangLearningHowActive2017,liuLearningHowActively2018},
and is made possible by the recent advances in pre-trained language models~\citep{devlinBERTPretrainingDeep2019,liuRoBERTaRobustlyOptimized2019,raffelExploringLimitsTransfer2020a,gaoMakingPretrainedLanguage2021}.
In-context learning, the observation that LMs \citep{radford2019language,
brownLanguageModelsAre2020, raeScalingLanguageModels2022,
zhangOPTOpenPretrained2022} can ``learn'' to perform a task when conditioned on a prompt.
\citet{xieExplanationIncontextLearning2022} explains the emergenece of
in-context learning by inferring the shared latent concept among
demonstration examples, while \citet{minRethinkingRoleDemonstrations2022}
finds the success of in-context learning is largely independent of access to
gold labels.

A variety of issues with in-context learning is discovered,
including surface form competition, the phenomenon that multiple words referring
to the same concept fighting for probability mass
\citep{holtzmanSurfaceFormCompetition2021},
and sensitivity of LMs due to changes in prompt
\citep{lesterPowerScaleParameterEfficient2021},
instruction \citep{mishraReframingInstructionalPrompts2022}, or
ordering of demonstration examples \citep{zhaoCalibrateUseImproving2021,
luFantasticallyOrderedPrompts2022}.
To optimize the performance of in-context learning, methods with varying
levels of granularity are proposed. Such methods include prompt tuning
\citep{lesterPowerScaleParameterEfficient2021, vuSPoTBetterFrozen2022,
wuIDPGInstanceDependentPrompt2022}, and instruction optimization
\citep{mishraReframingInstructionalPrompts2022,
kojimaLargeLanguageModels2022}. \citet{liuWhatMakesGood2021} approaches
the example selection problem by searching for nearest neighbors of test
examples in the embedding space, while \citet{rubinLearningRetrievePrompts2022}
uses a scoring LM for example retrieval.

\section{Discussion}
\label{sec:discussion}

Inspired by \citet{Pang+Lee:05a}, we adopt a Q\&A format to discuss the implications of our work.

{\bf Q:} Are GPT-2 results still relevant?

{\bf A:} We believe that it is relevant for three reasons.
First, GPT-2 is public and economically feasible options for many researchers.
Our knowledge about GPT-2 is far from complete and expanding this understanding is useful on its own.
Second, in the long term, it is unclear that everyone will have access to large models or that it is appropriate to use the largest model available in every use case.
Models of moderate sizes are likely still useful depending on the use case.
Third, it is important to highlight the emerging abilities over different sizes of language models.
By understanding the phase change, i.e., when emerging abilities happen, we will better understand the behavior of large-scale language models.

That said, one should caution against making generalizing claims based on results from GPT-2, because the results may not generalize to GPT-3 \citep{bowman-2022-dangers}.
This is why we present negative results from GPT-3.
Differing results between GPT-2 and GPT-3 or more generally models of different sizes will be a reality in NLP for a while.
It is important for the NLP community to collectively build knowledge about such differences and develop the future ecosystem of models.

{\bf Q:} Why did you not experiment with GPT-3-Davinci?

{\bf A:} The goal of this work is twofold: 1) assessing the ability of large-scale language models to acquire new information and 2) exploring whether reinforcement learning can identify reliable strategies for actively selecting examples.
Our results are generally positive on GPT-2.
Meanwhile, we observe relatively small variance after calibration with GPT-3-Babbage, so it does not seem economically sensible to experiment with even bigger models.

{\bf Q:} Why did you choose $k=4$? Is this generalizable?

{\bf A:} Our experiments are limited by the context window of GPT-2
(1024 tokens) and GPT-3 (2048) tokens.
Using $k$ beyond 4 would frequently leads to demonstration examples overflowing
the token limit and need to be truncated.
Additionally, prior work~\citep{zhaoCalibrateUseImproving2021,brownLanguageModelsAre2020}
shows diminishing improvements of in-context learning performance by adding
the number of demonstration examples beyond $4$.
Therefore, we believe experimenting with $k=4$ is a reasonable choice.
We are optimistic that our framework and method can generalize to different
shots.

\section{Conclusion}

In this work, we investigate how large language models acquire information through the perspective of example selection for in-context learning.
In-context learning with GPT-2 and GPT-3 is sensitive to the selection of demonstration examples.
In order to identify generalizable properties of useful demonstration examples, we study active example selection where unlabeled examples are iteratively selected, annotated, and added to the prompt.
We use reinforcement learning to train policies for active example selection.
The learned policy stablizes in-context learning and improves accuracy when we apply it to a new pool of unlabeled examples or even completely new tasks unseen during training for GPT-2.
Our analyses further reveal that properties of useful demonstration examples can deviate from human intuitions.

Examples selected from GPT-2 can still lead to a small improvement on GPT-3 Ada, however, the gain diminishes on larger models (i.e., Babbage and Curie).
Our results highlight the challenges of generalization in the era of large-scale models due to their emerging capabilities.
We believe that it is important for the NLP community to collectively build knowledge about such differences and develop the future ecosystem of models together.

\section*{Ethics Statement}

Our primary goal is to understand how large language models acquire new information in in-context learning through the perspective of example selection.
A better understanding can help develop more effective strategies for in-context learning as well as better large-scale language models.
However, these strategies can also be used in applications that may incur harm to the society.

\section*{Acknowledgments}

We thank all anonymous reviewers for their insightful suggestions and comments. We thank all members of the Chicago Human+AI Lab for feedback on early versions of this work. This work was supported in part by an Amazon research award, a Salesforce research award, a UChicago DSI discovery grant, and an NSF grant IIS-2126602.

\bibliography{auto,refs}

\begin{thebibliography}{44}
\expandafter\ifx\csname natexlab\endcsname\relax\def\natexlab#1{#1}\fi

\bibitem[{Arulkumaran et~al.(2017)Arulkumaran, Deisenroth, Brundage, and
  Bharath}]{arulkumaranBriefSurveyDeep2017}
Kai Arulkumaran, Marc~Peter Deisenroth, Miles Brundage, and Anil~Anthony
  Bharath. 2017.
\newblock \href {https://doi.org/10.1109/MSP.2017.2743240} {A {{Brief Survey}}
  of {{Deep Reinforcement Learning}}}.
\newblock \emph{IEEE Signal Processing Magazine}, 34(6):26--38.

\bibitem[{Bellman(1957)}]{bellmanDynamicProgramming1957}
Richard Bellman. 1957.
\newblock \emph{Dynamic Programming}, first edition.
\newblock {Princeton University Press}, {Princeton, NJ, USA}.

\bibitem[{Bowman(2022)}]{bowman-2022-dangers}
Samuel Bowman. 2022.
\newblock \href {https://doi.org/10.18653/v1/2022.acl-long.516} {The dangers of
  underclaiming: Reasons for caution when reporting how {NLP} systems fail}.
\newblock In \emph{Proceedings of the 60th Annual Meeting of the Association
  for Computational Linguistics (Volume 1: Long Papers)}, pages 7484--7499,
  Dublin, Ireland. Association for Computational Linguistics.

\bibitem[{Brown et~al.(2020)Brown, Mann, Ryder, Subbiah, Kaplan, Dhariwal,
  Neelakantan, Shyam, Sastry, Askell, Agarwal, {Herbert-Voss}, Krueger,
  Henighan, Child, Ramesh, Ziegler, Wu, Winter, Hesse, Chen, Sigler, Litwin,
  Gray, Chess, Clark, Berner, McCandlish, Radford, Sutskever, and
  Amodei}]{brownLanguageModelsAre2020}
Tom Brown, Benjamin Mann, Nick Ryder, Melanie Subbiah, Jared~D Kaplan, Prafulla
  Dhariwal, Arvind Neelakantan, Pranav Shyam, Girish Sastry, Amanda Askell,
  Sandhini Agarwal, Ariel {Herbert-Voss}, Gretchen Krueger, Tom Henighan, Rewon
  Child, Aditya Ramesh, Daniel Ziegler, Jeffrey Wu, Clemens Winter, Chris
  Hesse, Mark Chen, Eric Sigler, Mateusz Litwin, Scott Gray, Benjamin Chess,
  Jack Clark, Christopher Berner, Sam McCandlish, Alec Radford, Ilya Sutskever,
  and Dario Amodei. 2020.
\newblock Language {{Models}} are {{Few-Shot Learners}}.
\newblock In \emph{Advances in {{Neural Information Processing Systems}}},
  volume~33, pages 1877--1901. {Curran Associates, Inc.}

\bibitem[{Dagan and Engelson(1995)}]{daganCommitteebasedSamplingTraining1995}
Ido Dagan and Sean~P. Engelson. 1995.
\newblock Committee-based sampling for training probabilistic classifiers.
\newblock In \emph{Proceedings of the Twelfth International Conference on
  International Conference on Machine Learning}, {{ICML}}'95, pages 150--157,
  {San Francisco, CA, USA}. {Morgan Kaufmann Publishers Inc.}

\bibitem[{Devlin et~al.(2019)Devlin, Chang, Lee, and
  Toutanova}]{devlinBERTPretrainingDeep2019}
Jacob Devlin, Ming-Wei Chang, Kenton Lee, and Kristina Toutanova. 2019.
\newblock \href {https://doi.org/10.18653/v1/N19-1423} {{{BERT}}:
  {{Pre-training}} of {{Deep Bidirectional Transformers}} for {{Language
  Understanding}}}.
\newblock In \emph{Proceedings of the 2019 {{Conference}} of the {{North
  American Chapter}} of the {{Association}} for {{Computational Linguistics}}:
  {{Human Language Technologies}}, {{Volume}} 1 ({{Long}} and {{Short
  Papers}})}, pages 4171--4186, {Minneapolis, Minnesota}. {Association for
  Computational Linguistics}.

\bibitem[{Fang et~al.(2017)Fang, Li, and Cohn}]{fangLearningHowActive2017}
Meng Fang, Yuan Li, and Trevor Cohn. 2017.
\newblock \href {https://doi.org/10.18653/v1/D17-1063} {Learning how to
  {{Active Learn}}: {{A Deep Reinforcement Learning Approach}}}.
\newblock In \emph{Proceedings of the 2017 {{Conference}} on {{Empirical
  Methods}} in {{Natural Language Processing}}}, pages 595--605, {Copenhagen,
  Denmark}. {Association for Computational Linguistics}.

\bibitem[{Gao et~al.(2021)Gao, Fisch, and
  Chen}]{gaoMakingPretrainedLanguage2021}
Tianyu Gao, Adam Fisch, and Danqi Chen. 2021.
\newblock \href {https://doi.org/10.18653/v1/2021.acl-long.295} {Making
  {{Pre-trained Language Models Better Few-shot Learners}}}.
\newblock In \emph{Proceedings of the 59th {{Annual Meeting}} of the
  {{Association}} for {{Computational Linguistics}} and the 11th
  {{International Joint Conference}} on {{Natural Language Processing}}
  ({{Volume}} 1: {{Long Papers}})}, pages 3816--3830, {Online}. {Association
  for Computational Linguistics}.

\bibitem[{Hasselt(2010)}]{hasseltDoubleQlearning2010}
Hado Hasselt. 2010.
\newblock Double {{Q-learning}}.
\newblock In \emph{Advances in {{Neural Information Processing Systems}}},
  volume~23. {Curran Associates, Inc.}

\bibitem[{Hessel et~al.(2017)Hessel, Modayil, {van Hasselt}, Schaul, Ostrovski,
  Dabney, Horgan, Piot, Azar, and
  Silver}]{hesselRainbowCombiningImprovements2017}
Matteo Hessel, Joseph Modayil, Hado {van Hasselt}, Tom Schaul, Georg Ostrovski,
  Will Dabney, Dan Horgan, Bilal Piot, Mohammad Azar, and David Silver. 2017.
\newblock \href {http://arxiv.org/abs/1710.02298} {Rainbow: {{Combining
  Improvements}} in {{Deep Reinforcement Learning}}}.

\bibitem[{Holtzman et~al.(2021)Holtzman, West, Shwartz, Choi, and
  Zettlemoyer}]{holtzmanSurfaceFormCompetition2021}
Ari Holtzman, Peter West, Vered Shwartz, Yejin Choi, and Luke Zettlemoyer.
  2021.
\newblock \href {http://arxiv.org/abs/2104.08315} {Surface {{Form
  Competition}}: {{Why}} the {{Highest Probability Answer Isn}}'t {{Always
  Right}}}.

\bibitem[{Kirk et~al.(2022)Kirk, Zhang, Grefenstette, and
  Rockt{\"a}schel}]{kirkSurveyGeneralisationDeep2022}
Robert Kirk, Amy Zhang, Edward Grefenstette, and Tim Rockt{\"a}schel. 2022.
\newblock \href {http://arxiv.org/abs/2111.09794} {A {{Survey}} of
  {{Generalisation}} in {{Deep Reinforcement Learning}}}.

\bibitem[{Kojima et~al.(2022)Kojima, Gu, Reid, Matsuo, and
  Iwasawa}]{kojimaLargeLanguageModels2022}
Takeshi Kojima, Shixiang~Shane Gu, Machel Reid, Yutaka Matsuo, and Yusuke
  Iwasawa. 2022.
\newblock \href {https://doi.org/10.48550/arXiv.2205.11916} {Large {{Language
  Models}} are {{Zero-Shot Reasoners}}}.

\bibitem[{Kumar et~al.(2020)Kumar, Zhou, Tucker, and
  Levine}]{kumarConservativeQLearningOffline2020}
Aviral Kumar, Aurick Zhou, George Tucker, and Sergey Levine. 2020.
\newblock \href {http://arxiv.org/abs/2006.04779} {Conservative {{Q-Learning}}
  for {{Offline Reinforcement Learning}}}.

\bibitem[{Lester et~al.(2021)Lester, {Al-Rfou}, and
  Constant}]{lesterPowerScaleParameterEfficient2021}
Brian Lester, Rami {Al-Rfou}, and Noah Constant. 2021.
\newblock \href {http://arxiv.org/abs/2104.08691} {The {{Power}} of {{Scale}}
  for {{Parameter-Efficient Prompt Tuning}}}.
\newblock \emph{arXiv:2104.08691 [cs]}.

\bibitem[{Lin(1992)}]{linSelfImprovingReactiveAgents1992}
Long-Ji Lin. 1992.
\newblock \href {https://doi.org/10.1007/BF00992699} {Self-{{Improving Reactive
  Agents Based}} on {{Reinforcement Learning}}, {{Planning}} and {{Teaching}}}.
\newblock \emph{Machine Language}, 8(3-4):293--321.

\bibitem[{Liu et~al.(2021)Liu, Shen, Zhang, Dolan, Carin, and
  Chen}]{liuWhatMakesGood2021}
Jiachang Liu, Dinghan Shen, Yizhe Zhang, Bill Dolan, Lawrence Carin, and Weizhu
  Chen. 2021.
\newblock \href {http://arxiv.org/abs/2101.06804} {What {{Makes Good In-Context
  Examples}} for {{GPT-}}\$3\$?}

\bibitem[{Liu et~al.(2018)Liu, Buntine, and
  Haffari}]{liuLearningHowActively2018}
Ming Liu, Wray Buntine, and Gholamreza Haffari. 2018.
\newblock \href {https://doi.org/10.18653/v1/P18-1174} {Learning {{How}} to
  {{Actively Learn}}: {{A Deep Imitation Learning Approach}}}.
\newblock In \emph{Proceedings of the 56th {{Annual Meeting}} of the
  {{Association}} for {{Computational Linguistics}} ({{Volume}} 1: {{Long
  Papers}})}, pages 1874--1883, {Melbourne, Australia}. {Association for
  Computational Linguistics}.

\bibitem[{Liu et~al.(2019)Liu, Ott, Goyal, Du, Joshi, Chen, Levy, Lewis,
  Zettlemoyer, and Stoyanov}]{liuRoBERTaRobustlyOptimized2019}
Yinhan Liu, Myle Ott, Naman Goyal, Jingfei Du, Mandar Joshi, Danqi Chen, Omer
  Levy, Mike Lewis, Luke Zettlemoyer, and Veselin Stoyanov. 2019.
\newblock \href {http://arxiv.org/abs/1907.11692} {{{RoBERTa}}: {{A Robustly
  Optimized BERT Pretraining Approach}}}.

\bibitem[{Lu et~al.(2022)Lu, Bartolo, Moore, Riedel, and
  Stenetorp}]{luFantasticallyOrderedPrompts2022}
Yao Lu, Max Bartolo, Alastair Moore, Sebastian Riedel, and Pontus Stenetorp.
  2022.
\newblock \href {https://doi.org/10.18653/v1/2022.acl-long.556} {Fantastically
  {{Ordered Prompts}} and {{Where}} to {{Find Them}}: {{Overcoming Few-Shot
  Prompt Order Sensitivity}}}.
\newblock In \emph{Proceedings of the 60th {{Annual Meeting}} of the
  {{Association}} for {{Computational Linguistics}} ({{Volume}} 1: {{Long
  Papers}})}, pages 8086--8098, {Dublin, Ireland}. {Association for
  Computational Linguistics}.

\bibitem[{Min et~al.(2022)Min, Lyu, Holtzman, Artetxe, Lewis, Hajishirzi, and
  Zettlemoyer}]{minRethinkingRoleDemonstrations2022}
Sewon Min, Xinxi Lyu, Ari Holtzman, Mikel Artetxe, Mike Lewis, Hannaneh
  Hajishirzi, and Luke Zettlemoyer. 2022.
\newblock \href {http://arxiv.org/abs/2202.12837} {Rethinking the {{Role}} of
  {{Demonstrations}}: {{What Makes In-Context Learning Work}}?}
\newblock \emph{arXiv:2202.12837 [cs]}.

\bibitem[{Mishra et~al.(2022)Mishra, Khashabi, Baral, Choi, and
  Hajishirzi}]{mishraReframingInstructionalPrompts2022}
Swaroop Mishra, Daniel Khashabi, Chitta Baral, Yejin Choi, and Hannaneh
  Hajishirzi. 2022.
\newblock \href {http://arxiv.org/abs/2109.07830} {Reframing {{Instructional
  Prompts}} to {{GPTk}}'s {{Language}}}.

\bibitem[{Mnih et~al.(2013)Mnih, Kavukcuoglu, Silver, Graves, Antonoglou,
  Wierstra, and Riedmiller}]{mnihPlayingAtariDeep2013}
Volodymyr Mnih, Koray Kavukcuoglu, David Silver, Alex Graves, Ioannis
  Antonoglou, Daan Wierstra, and Martin Riedmiller. 2013.
\newblock \href {https://doi.org/10.48550/arXiv.1312.5602} {Playing {{Atari}}
  with {{Deep Reinforcement Learning}}}.

\bibitem[{Nakano et~al.(2022)Nakano, Hilton, Balaji, Wu, Ouyang, Kim, Hesse,
  Jain, Kosaraju, Saunders, Jiang, Cobbe, Eloundou, Krueger, Button, Knight,
  Chess, and Schulman}]{nakanoWebGPTBrowserassistedQuestionanswering2022}
Reiichiro Nakano, Jacob Hilton, Suchir Balaji, Jeff Wu, Long Ouyang, Christina
  Kim, Christopher Hesse, Shantanu Jain, Vineet Kosaraju, William Saunders,
  Xu~Jiang, Karl Cobbe, Tyna Eloundou, Gretchen Krueger, Kevin Button, Matthew
  Knight, Benjamin Chess, and John Schulman. 2022.
\newblock \href {https://doi.org/10.48550/arXiv.2112.09332} {{{WebGPT}}:
  {{Browser-assisted}} question-answering with human feedback}.

\bibitem[{Ng et~al.(1999)Ng, Harada, and
  Russell}]{ngPolicyInvarianceReward1999}
Andrew~Y. Ng, Daishi Harada, and Stuart~J. Russell. 1999.
\newblock Policy {{Invariance Under Reward Transformations}}: {{Theory}} and
  {{Application}} to {{Reward Shaping}}.
\newblock In \emph{Proceedings of the {{Sixteenth International Conference}} on
  {{Machine Learning}}}, {{ICML}} '99, pages 278--287, {San Francisco, CA,
  USA}. {Morgan Kaufmann Publishers Inc.}

\bibitem[{Pang and Lee(2005)}]{Pang+Lee:05a}
Bo~Pang and Lillian Lee. 2005.
\newblock Seeing stars: Exploiting class relationships for sentiment
  categorization with respect to rating scales.
\newblock In \emph{Proceedings of ACL}, pages 115--124.

\bibitem[{Pathak et~al.(2017)Pathak, Agrawal, Efros, and
  Darrell}]{pathakCuriosityDrivenExplorationSelfSupervised2017}
Deepak Pathak, Pulkit Agrawal, Alexei~A. Efros, and Trevor Darrell. 2017.
\newblock \href {https://doi.org/10.1109/CVPRW.2017.70} {Curiosity-{{Driven
  Exploration}} by {{Self-Supervised Prediction}}}.
\newblock In \emph{2017 {{IEEE Conference}} on {{Computer Vision}} and
  {{Pattern Recognition Workshops}} ({{CVPRW}})}, pages 488--489, {Honolulu,
  HI, USA}. {IEEE}.

\bibitem[{Perez et~al.(2021)Perez, Kiela, and
  Cho}]{perezTrueFewShotLearning2021}
Ethan Perez, Douwe Kiela, and Kyunghyun Cho. 2021.
\newblock True {{Few-Shot Learning}} with {{Language Models}}.
\newblock In \emph{Advances in {{Neural Information Processing Systems}}}.

\bibitem[{Prudencio et~al.(2022)Prudencio, Maximo, and
  Colombini}]{prudencioSurveyOfflineReinforcement2022}
Rafael~Figueiredo Prudencio, Marcos R. O.~A. Maximo, and Esther~Luna Colombini.
  2022.
\newblock \href {http://arxiv.org/abs/2203.01387} {A {{Survey}} on {{Offline
  Reinforcement Learning}}: {{Taxonomy}}, {{Review}}, and {{Open Problems}}}.

\bibitem[{Radford et~al.(2019)Radford, Wu, Child, Luan, Amodei, and
  Sutskever}]{radford2019language}
Alec Radford, Jeff Wu, Rewon Child, David Luan, Dario Amodei, and Ilya
  Sutskever. 2019.
\newblock Language models are unsupervised multitask learners.

\bibitem[{Rae et~al.(2022)Rae, Borgeaud, Cai, Millican, Hoffmann, Song,
  Aslanides, Henderson, Ring, Young, Rutherford, Hennigan, Menick, Cassirer,
  Powell, van~den Driessche, Hendricks, Rauh, Huang, Glaese, Welbl, Dathathri,
  Huang, Uesato, Mellor, Higgins, Creswell, McAleese, Wu, Elsen, Jayakumar,
  Buchatskaya, Budden, Sutherland, Simonyan, Paganini, Sifre, Martens, Li,
  Kuncoro, Nematzadeh, Gribovskaya, Donato, Lazaridou, Mensch, Lespiau,
  Tsimpoukelli, Grigorev, Fritz, Sottiaux, Pajarskas, Pohlen, Gong, Toyama,
  {d'Autume}, Li, Terzi, Mikulik, Babuschkin, Clark, Casas, Guy, Jones,
  Bradbury, Johnson, Hechtman, Weidinger, Gabriel, Isaac, Lockhart, Osindero,
  Rimell, Dyer, Vinyals, Ayoub, Stanway, Bennett, Hassabis, Kavukcuoglu, and
  Irving}]{raeScalingLanguageModels2022}
Jack~W. Rae, Sebastian Borgeaud, Trevor Cai, Katie Millican, Jordan Hoffmann,
  Francis Song, John Aslanides, Sarah Henderson, Roman Ring, Susannah Young,
  Eliza Rutherford, Tom Hennigan, Jacob Menick, Albin Cassirer, Richard Powell,
  George van~den Driessche, Lisa~Anne Hendricks, Maribeth Rauh, Po-Sen Huang,
  Amelia Glaese, Johannes Welbl, Sumanth Dathathri, Saffron Huang, Jonathan
  Uesato, John Mellor, Irina Higgins, Antonia Creswell, Nat McAleese, Amy Wu,
  Erich Elsen, Siddhant Jayakumar, Elena Buchatskaya, David Budden, Esme
  Sutherland, Karen Simonyan, Michela Paganini, Laurent Sifre, Lena Martens,
  Xiang~Lorraine Li, Adhiguna Kuncoro, Aida Nematzadeh, Elena Gribovskaya,
  Domenic Donato, Angeliki Lazaridou, Arthur Mensch, Jean-Baptiste Lespiau,
  Maria Tsimpoukelli, Nikolai Grigorev, Doug Fritz, Thibault Sottiaux, Mantas
  Pajarskas, Toby Pohlen, Zhitao Gong, Daniel Toyama, Cyprien de~Masson
  {d'Autume}, Yujia Li, Tayfun Terzi, Vladimir Mikulik, Igor Babuschkin, Aidan
  Clark, Diego de~Las Casas, Aurelia Guy, Chris Jones, James Bradbury, Matthew
  Johnson, Blake Hechtman, Laura Weidinger, Iason Gabriel, William Isaac,
  Ed~Lockhart, Simon Osindero, Laura Rimell, Chris Dyer, Oriol Vinyals, Kareem
  Ayoub, Jeff Stanway, Lorrayne Bennett, Demis Hassabis, Koray Kavukcuoglu, and
  Geoffrey Irving. 2022.
\newblock \href {http://arxiv.org/abs/2112.11446} {Scaling {{Language Models}}:
  {{Methods}}, {{Analysis}} \& {{Insights}} from {{Training Gopher}}}.

\bibitem[{Raffel et~al.(2020)Raffel, Shazeer, Roberts, Lee, Narang, Matena,
  Zhou, Li, and Liu}]{raffelExploringLimitsTransfer2020a}
Colin Raffel, Noam Shazeer, Adam Roberts, Katherine Lee, Sharan Narang, Michael
  Matena, Yanqi Zhou, Wei Li, and Peter~J. Liu. 2020.
\newblock \href {http://arxiv.org/abs/1910.10683} {Exploring the {{Limits}} of
  {{Transfer Learning}} with a {{Unified Text-to-Text Transformer}}}.

\bibitem[{Rubin et~al.(2022)Rubin, Herzig, and
  Berant}]{rubinLearningRetrievePrompts2022}
Ohad Rubin, Jonathan Herzig, and Jonathan Berant. 2022.
\newblock \href {https://doi.org/10.48550/arXiv.2112.08633} {Learning {{To
  Retrieve Prompts}} for {{In-Context Learning}}}.

\bibitem[{Settles(2009)}]{settlesActiveLearningLiterature2009}
Burr Settles. 2009.
\newblock Active learning literature survey.

\bibitem[{Socher et~al.(2013)Socher, Perelygin, Wu, Chuang, Manning, Ng, and
  Potts}]{socherRecursiveDeepModels2013}
Richard Socher, Alex Perelygin, Jean Wu, Jason Chuang, Christopher~D. Manning,
  Andrew Ng, and Christopher Potts. 2013.
\newblock Recursive {{Deep Models}} for {{Semantic Compositionality Over}} a
  {{Sentiment Treebank}}.
\newblock In \emph{Proceedings of the 2013 {{Conference}} on {{Empirical
  Methods}} in {{Natural Language Processing}}}, pages 1631--1642, {Seattle,
  Washington, USA}. {Association for Computational Linguistics}.

\bibitem[{Von~Wieser(1893)}]{von1893natural}
Friedrich~Freiherr Von~Wieser. 1893.
\newblock \emph{Natural Value}.
\newblock {Macmillan and Company}.

\bibitem[{Voorhees and Tice(2000)}]{voorheesBuildingQuestionAnswering2000}
Ellen~M. Voorhees and Dawn~M. Tice. 2000.
\newblock \href {https://doi.org/10.1145/345508.345577} {Building a question
  answering test collection}.
\newblock In \emph{Proceedings of the 23rd Annual International {{ACM SIGIR}}
  Conference on {{Research}} and Development in Information Retrieval},
  {{SIGIR}} '00, pages 200--207, {New York, NY, USA}. {Association for
  Computing Machinery}.

\bibitem[{Vu et~al.(2022)Vu, Lester, Constant, {Al-Rfou}, and
  Cer}]{vuSPoTBetterFrozen2022}
Tu~Vu, Brian Lester, Noah Constant, Rami {Al-Rfou}, and Daniel Cer. 2022.
\newblock \href {http://arxiv.org/abs/2110.07904} {{{SPoT}}: {{Better Frozen
  Model Adaptation}} through {{Soft Prompt Transfer}}}.
\newblock \emph{arXiv:2110.07904 [cs]}.

\bibitem[{Wei et~al.(2022)Wei, Tay, Bommasani, Raffel, Zoph, Borgeaud,
  Yogatama, Bosma, Zhou, Metzler et~al.}]{wei2022emergent}
Jason Wei, Yi~Tay, Rishi Bommasani, Colin Raffel, Barret Zoph, Sebastian
  Borgeaud, Dani Yogatama, Maarten Bosma, Denny Zhou, Donald Metzler, et~al.
  2022.
\newblock Emergent abilities of large language models.
\newblock \emph{arXiv preprint arXiv:2206.07682}.

\bibitem[{Wu et~al.(2022)Wu, Wang, Gu, Hou, Dong, Vydiswaran, and
  Ma}]{wuIDPGInstanceDependentPrompt2022}
Zhuofeng Wu, Sinong Wang, Jiatao Gu, Rui Hou, Yuxiao Dong, V.~G.~Vinod
  Vydiswaran, and Hao Ma. 2022.
\newblock \href {http://arxiv.org/abs/2204.04497} {{{IDPG}}: {{An
  Instance-Dependent Prompt Generation Method}}}.
\newblock \emph{arXiv:2204.04497 [cs]}.

\bibitem[{Xie et~al.(2022)Xie, Raghunathan, Liang, and
  Ma}]{xieExplanationIncontextLearning2022}
Sang~Michael Xie, Aditi Raghunathan, Percy Liang, and Tengyu Ma. 2022.
\newblock \href {http://arxiv.org/abs/2111.02080} {An {{Explanation}} of
  {{In-context Learning}} as {{Implicit Bayesian Inference}}}.

\bibitem[{Zhang et~al.(2022)Zhang, Roller, Goyal, Artetxe, Chen, Chen, Dewan,
  Diab, Li, Lin, Mihaylov, Ott, Shleifer, Shuster, Simig, Koura, Sridhar, Wang,
  and Zettlemoyer}]{zhangOPTOpenPretrained2022}
Susan Zhang, Stephen Roller, Naman Goyal, Mikel Artetxe, Moya Chen, Shuohui
  Chen, Christopher Dewan, Mona Diab, Xian Li, Xi~Victoria Lin, Todor Mihaylov,
  Myle Ott, Sam Shleifer, Kurt Shuster, Daniel Simig, Punit~Singh Koura, Anjali
  Sridhar, Tianlu Wang, and Luke Zettlemoyer. 2022.
\newblock \href {https://doi.org/10.48550/arXiv.2205.01068} {{{OPT}}: {{Open
  Pre-trained Transformer Language Models}}}.

\bibitem[{Zhang et~al.(2015)Zhang, Zhao, and
  LeCun}]{zhangCharacterlevelConvolutionalNetworks2015}
Xiang Zhang, Junbo Zhao, and Yann LeCun. 2015.
\newblock Character-level {{Convolutional Networks}} for {{Text
  Classification}}.
\newblock In \emph{Advances in {{Neural Information Processing Systems}}},
  volume~28. {Curran Associates, Inc.}

\bibitem[{Zhao et~al.(2021)Zhao, Wallace, Feng, Klein, and
  Singh}]{zhaoCalibrateUseImproving2021}
Tony~Z. Zhao, Eric Wallace, Shi Feng, Dan Klein, and Sameer Singh. 2021.
\newblock \href {http://arxiv.org/abs/2102.09690} {Calibrate {{Before Use}}:
  {{Improving Few-Shot Performance}} of {{Language Models}}}.

\end{thebibliography}
\bibliographystyle{acl_natbib}

\appendix

\section{Conservative Q-Learning}
\label{appendix:cql}

The objective of standard Q-learning is to minimize the Bellman Error (BE):

\begin{multline*}
    \text{BE}(Q) = \mathbb{E}_{s, a, s' \sim \mathcal{D}}
    \Big[ \reward(s, \action) + \\
    \gamma \max_{\action'} Q(s', \action') - Q(s, a) \Big] \text{.}
\end{multline*}

An issue with offline Q-learning is there are OOD actions that do not appear in
 the training data.
Learned Q-networks often overestimate these Q-values, resulting in the policy
 taking unfamiliar actions during evaluation and hurts performance.
To mitigate this issue, conservative Q-learning (CQL) adds a
 penalty term to regularize Q-values:

\begin{multline*}
    \min_Q \alpha \mathbb{E}_{s \sim \mathcal{D}} \Big[ \log \sum_a \exp(Q(s, a)) - \\ 
    \mathbb{E}_{a \sim \hat{\pi}_\beta}\big[Q(s, a)\big] \Big] + \frac{1}{2} \text{BE}(Q)^2 \text{,}
\end{multline*}
where $\alpha$ is a weight term, and $\hat{\pi}_\beta$ is the {\em behavior policy},
under which the offline transitions are collected for training.
Notice this objective penalizes all unobserved actions under $\hat{\pi}_\beta$.
Intuitively, this regularizer leads to a policy that avoids unfamiliar
actions during evaluation.
We refer the interested reader to the original paper for theoretical guarantees
and further details~\citep{kumarConservativeQLearningOffline2020}.

\section{Hyperparameters}
\label{appendix:hyperparameters}

\begin{table}[h]
	\centering
	\begin{tabular}{@{}l|l@{}}
		\toprule
		Hyperparameter                     & Value            \\
		\midrule
		Train steps                        & 8000             \\
		Batch size                         & 16               \\
		Hidden dim (MLP)                   & 16               \\
		Replay memory size                 & 50000            \\
		Learning rate                      & 1e-4, 3e-4, 5e-4 \\
		CQL regularization weight $\alpha$ & 0, 0.1, 0.2      \\
		Target network update steps        & 100, 200, 400    \\
		Dropout rate                       & 0, 0.25          \\
		\bottomrule
	\end{tabular}
	\caption{List of hyperparameters used in our experiments.}
	\label{tab:hyper}
\end{table}

We report the list of hyperparameters for the hyperparameter search in
 Table~\ref{tab:hyper}.
We use grid search over these hyperparameters to determine the combination that
 maximizes validation performance.
During validation, the policy picks from the \textbf{reward set}, and is
 evaluated on the \textbf{training set}, whereas in training, we pick from the
 \textbf{training set} and evaluate on the \textbf{reward set}.
We point out that our validation scheme does not use extra data.

\begin{table*}[t]
    \centering
    \begin{tabular}{@{}lccccc@{}}
    \toprule
    Method                                  & Average   & \agnews         &\amazon        & \sst          & \trec         \\ \midrule
    random                                  & $59.6$    & $55.2_{10.5}$   & $76.3_{12.3}$ & $66.2_{12.9}$ & $40.8_{4.7}$  \\
    max-entropy                             & $59.3$    & $58.8_{11.3}$   & $74.8_{5.1}$  & $65.7_{10.7}$ & $37.8_{6.7}$  \\ \midrule
    best-of-10                              & $72.5$    & $72.1_{1.9}$    & $91.1_{0.6}$  & $81.1_{4.4}$  & $45.6_{3.5}$  \\
    greedy-oracle                           & $78.0$    & $80.6_{1.7}$    & $91.8_{1.1}$  & $81.7_{3.9}$  & $58.0_{7.5}$  \\ \midrule
    Linear policy ({\bf seen examples})     & $65.6$    & $62.8_{7.8}$    & $82.7_{8.6}$  & $74.2_{5.8}$  & $42.8_{2.9}$  \\
    Linear policy ({\bf 1000 new examples}) & $65.9$    & $69.5_{6.0}$    & $83.7_{6.2}$  & $65.2_{4.9}$  & $45.2_{2.8}$  \\ \midrule
    MLP policy ({\bf seen examples})        & $71.4$    & $70.8_{7.8}$    & $90.4_{1.9}$  & $81.0_{3.5}$  & $43.3_{2.0}$  \\
    MLP policy ({\bf 1000 new examples})    & $69.0$    & $65.5_{7.4}$    & $88.5_{4.2}$  & $76.7_{7.5}$  & $45.4_{5.0}$  \\ \bottomrule
    \end{tabular}
    \caption{{\bf \sametask} accuracy on \agnews, \amazon, \sst and \trec,
    across 5 random seeds, with our methods (using MLP and Linear networks as
    policies).
    $95\%$ confidence intervals are reported as subscripts.}
    \label{table:linear}
\end{table*}

Table~\ref{table:linear} further includes the performance of linear policies.
The performance of linear policies is better than the baselines, but clearly
 worse than the MLP policy.

\section{Additional Results}

We present results on the effect of unlabeled size and on transfer GPT-3.
We also provide additional analysis towards understanding what makes good examples for
 in-context learning.

\begin{figure}[t]
	\centering

	\includegraphics[width=0.48\textwidth]{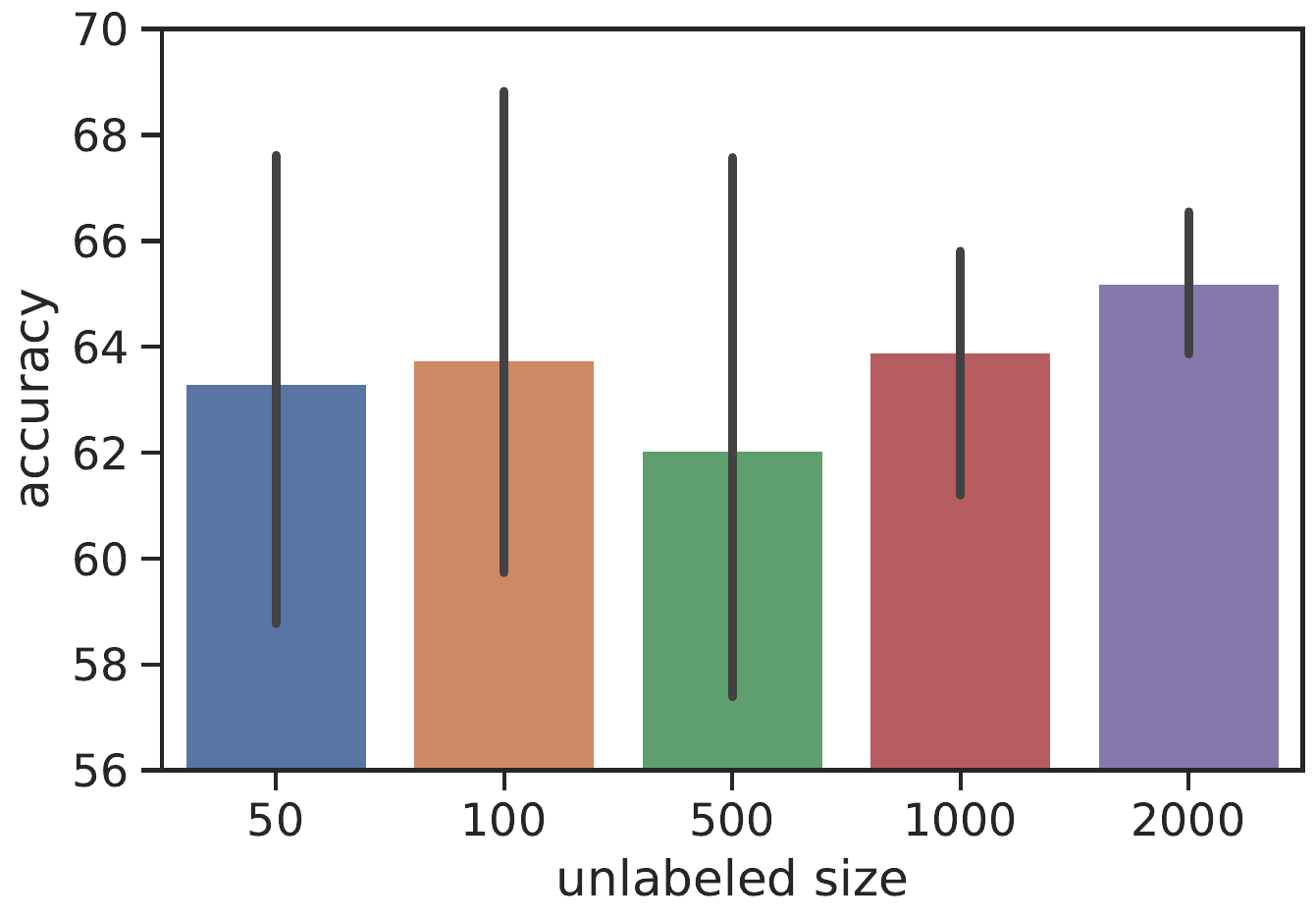}
	\caption{Average \newtask (transfer) accuracy on 4 tasks across 5 random
		seeds.
		$95\%$ confidence intervals are reported as error bars.}
	\label{fig:unlabeled-size}
\end{figure}

\subsection{Effect of Unlabeled Size}
\label{sec:unlabeled-size}

In \secref{sec:main-results}, we noticed the number of unlabeled examples
 available for selection plays a role in the performance our policies.
One might expect the transfer performance in the \newtask setting scales with
 unlabeled size, simply because there are additional examples to pick from.

In Figure~\ref{fig:unlabeled-size}, we plot average accuracies in the
 \newtask setting, where we train our policies on three datasets and evaluate on
 a held-out dataset.
Here, we notice the benefit of a larger unlabeled set is twofold, both in
 increasing transfer performance, and in reducing variance.
That said, the improvement is not necessarily monotonic due to the
 large variance.
Interestingly, our learned policy is performant even when the unlabeled set is
 small.
Picking from 50 unlabeled examples, our policies reaches an average accuracy of
 $63.3\%$, still manage to outperform random demonstration ($59.6\%$).

\subsection{Transfer to GPT-3}
\label{sec:gpt3-transfer}
\begin{table*}[t]
	\centering
	\begin{tabular}{@{}lccccc@{}}
		\toprule
		Model                                  & Average & \agnews              & \amazon             & \sst                 & \trec               \\ \midrule
		\ada                                   & $59.0$  & \statstd{62.9}{15.3} & \statstd{87.0}{5.3} & \statstd{65.0}{8.9}  & \statstd{21.2}{5.8} \\
		\ada ({\bf C})                         & $62.5$  & \statstd{64.0}{3.5}  & \statstd{90.0}{1.1} & \statstd{73.8}{8.5}  & \statstd{22.1}{4.6} \\
		\gpt policy $\to$ \ada                 & $60.1$  & \statstd{51.8}{15.5} & \statstd{89.1}{1.7} & \statstd{73.3}{15.0} & \statstd{26.2}{3.9} \\
		\gpt policy $\to$ \ada ({\bf C})       & $62.9$  & \statstd{55.6}{5.9}  & \statstd{89.7}{2.2} & \statstd{86.7}{1.6}  & \statstd{19.5}{1.4} \\
		\gpt examples $\to$ \ada               & $59.9$  & \statstd{48.9}{12.5} & \statstd{89.3}{2.5} & \statstd{74.8}{11.4} & \statstd{26.6}{3.9} \\
		\gpt examples $\to$ \ada ({\bf C})     & $64.4$  & \statstd{62.0}{8.3}  & \statstd{88.7}{3.2} & \statstd{84.0}{3.6}  & \statstd{23.0}{5.3} \\ \midrule
		\babbage                               & $70.3$  & \statstd{68.0}{12.3} & \statstd{93.4}{0.7} & \statstd{92.2}{2.4}  & \statstd{27.4}{5.1} \\
		\babbage ({\bf C})                     & $74.4$  & \statstd{78.1}{5.3}  & \statstd{92.7}{1.4} & \statstd{90.8}{1.0}  & \statstd{36.0}{3.5} \\
		\gpt policy $\to$ \babbage             & $68.7$  & \statstd{58.0}{5.9}  & \statstd{93.6}{2.2} & \statstd{90.6}{1.6}  & \statstd{32.5}{1.4} \\
		\gpt policy $\to$ \babbage ({\bf C})   & $74.4$  & \statstd{75.1}{5.3}  & \statstd{93.4}{0.5} & \statstd{90.3}{1.7}  & \statstd{38.8}{6.1} \\
		\gpt examples $\to$ \babbage           & $65.8$  & \statstd{42.6}{10.0} & \statstd{93.0}{0.4} & \statstd{91.1}{2.9}  & \statstd{36.6}{8.4} \\
		\gpt examples $\to$ \babbage ({\bf C}) & $73.6$  & \statstd{73.9}{7.3}  & \statstd{93.1}{0.5} & \statstd{91.1}{1.8}  & \statstd{36.2}{2.6} \\ \midrule
		\curie                                 & $74.2$  & $76.7$               & $94.7$              & $93.8$               & $31.4$              \\
		\curie ({\bf C})                       & $76.3$  & $69.8$               & $94.8$              & $93.4$               & $47.0$              \\
		\gpt policy $\to$ \curie               & $76.0$  & $81.2$               & $95.7$              & $96.0$               & $31.0$              \\
		\gpt policy $\to$ \curie ({\bf C})     & $75.4$  & $75.8$               & $95.4$              & $93.0$               & $38.2$              \\
		\gpt examples $\to$ \curie             & $74.4$  & $77.7$               & $93.8$              & $94.3$               & $31.8$              \\
		\gpt examples $\to$ \curie ({\bf C})   & $77.3$  & $79.8$               & $93.1$              & $94.6$               & $41.8$              \\
		\bottomrule
	\end{tabular}
	\caption{Transfer of policies and examples learned on \gpt to various GPT-3 models
		across 5 random sets of 4-shot demonstration examples. {\bf C} indicates
		calibration. $95\%$ confidence intervals are reported as subscripts.
		Due to resource constraints, we limit experiments with \curie to 1 random set.}
	\label{table:gpt3-transfer}
\end{table*}

Despite demonstrating abilities to generalize across tasks, it is yet clear
 whether learned policies on GPT-2 can generalize to other models, such as
 GPT-3.
In table~\ref{table:gpt3-transfer}, we report the performance of transferring
 both learned policies and selected examples from \gpt to GPT-3 \ada, \babbage and
 \curie.

We observe mixed results when transferring to GPT-3.
With an uncalibrated \ada model, we observe a small, but measurable improvement
 of transferring either policy ($1.1\%$) or examples directly ($0.9\%$).
Such a trend holds for the calibrated \ada model too ($0.4\%$ and $1.9\%$).
Despite the improved performance, the benefits of variance reduction is
 diminished.
Perhaps surprising is the generalization of learned policies: it suggests
 different models could indeed share similar preferences for demonstration
 examples.

On the other hand, we observe negative results when transferring to \babbage.
When transferring learned policy to an uncalibrated \babbage model, we notice
 the performance drops by $1.6\%$.
For cost considerations, we run \curie experiments for one random set and do not report variance.
Marginal gains are observed when transferring policy to the
 uncalibrated model ($1.8\%$) and examples to the calibrated model
 ($1.0\%$).
In other scenarios, transfer results match or underperform base models.
As the observed results could be attributed to randomness, we hold short of
 drawing conclusions.

\subsection{Coefficients in Linear Policies}
\label{appendix:linear}

\begin{figure*}[t]
	\centering
	\begin{subfigure}[b]{0.509\textwidth}
		\centering
		\includegraphics[width=\textwidth]{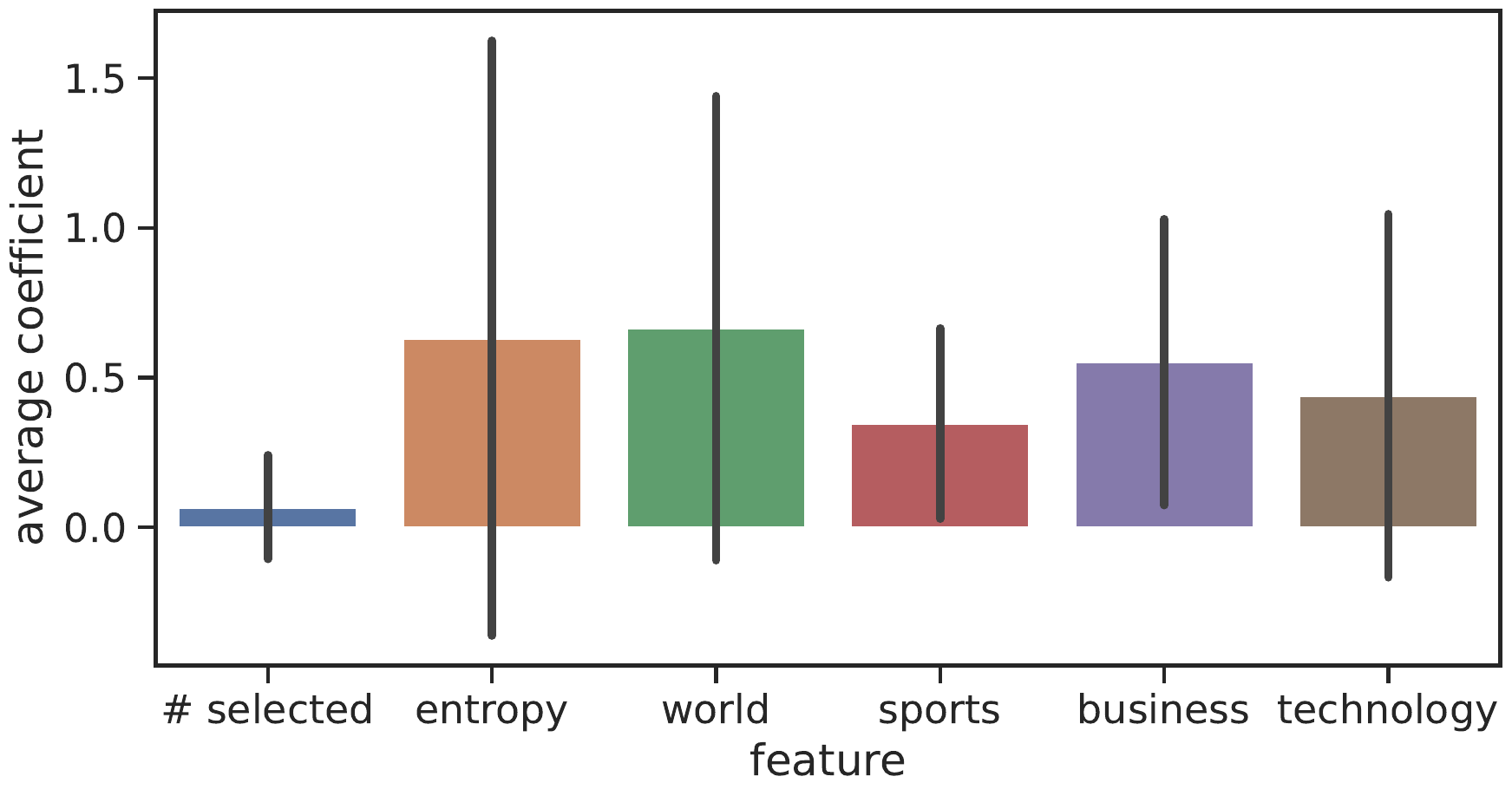}
		\caption{\agnews}
	\end{subfigure}
	\begin{subfigure}[b]{0.4\textwidth}
		\centering
		\includegraphics[width=\textwidth]{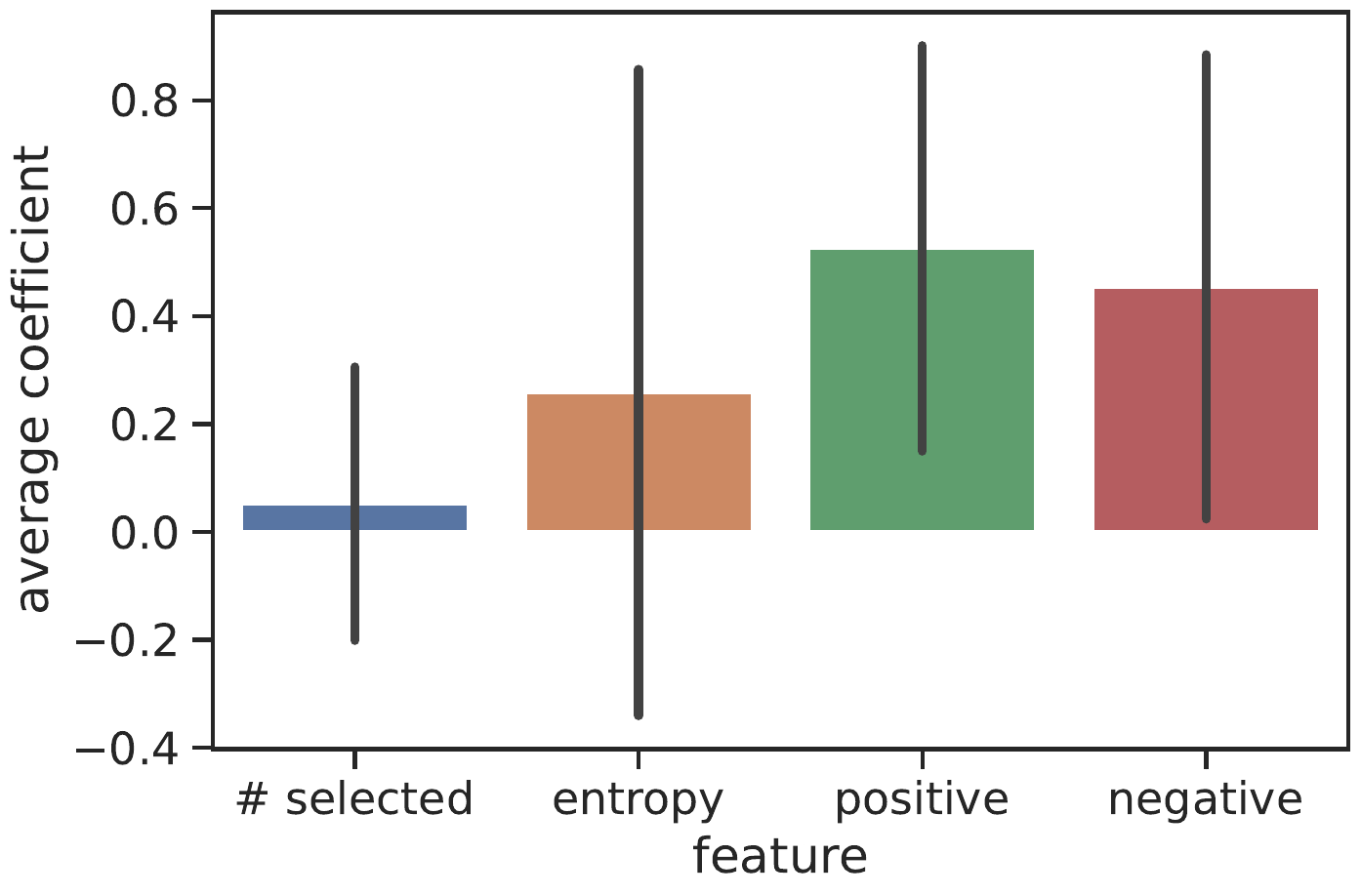}
		\caption{\sst}
	\end{subfigure}
	\caption{Average coefficients of linear policies trained on \agnews and
		\sst
		across 5 runs.
		Error bars show the standard deviation.
	}
	\label{fig:linear}
\end{figure*}

Although linear policies perform worse than the MLP, they are more
 interpretable.
Figure~\ref{fig:linear} shows the coefficients of feature representations of
 actions for \agnews and \sst.
The average coefficient of entropy is indeed positive, suggesting that
 strategies encouraging class balance have some value.
However, it is often not the most important feature.
For example, positive examples in \sst matter more, which is consistent with
 our observation in the main paper.
Moreover, the variance is large, highlighting the challenges in learning a
 generalizable policy.

\subsection{Effect of Length}
\label{appendix:length}

We also examine the effect of length on in-context learning.
Intuitively, one might expect longer examples to be more meaningful.
However, we do not see a correlation between length and accuracy in \agnews and
 \trec, and a non-significant negative correlations in \sst.
In \amazon, we observe a statistically significant (p-value = $0.019$), but
 weak correlation between length and accuracy.
Overall, there is no evidence suggesting longer examples improve in-context
 learning performance.

\begin{figure*}[t]
	\centering
	\begin{subfigure}[b]{0.4\textwidth}
		\centering

		\includegraphics[width=\textwidth]{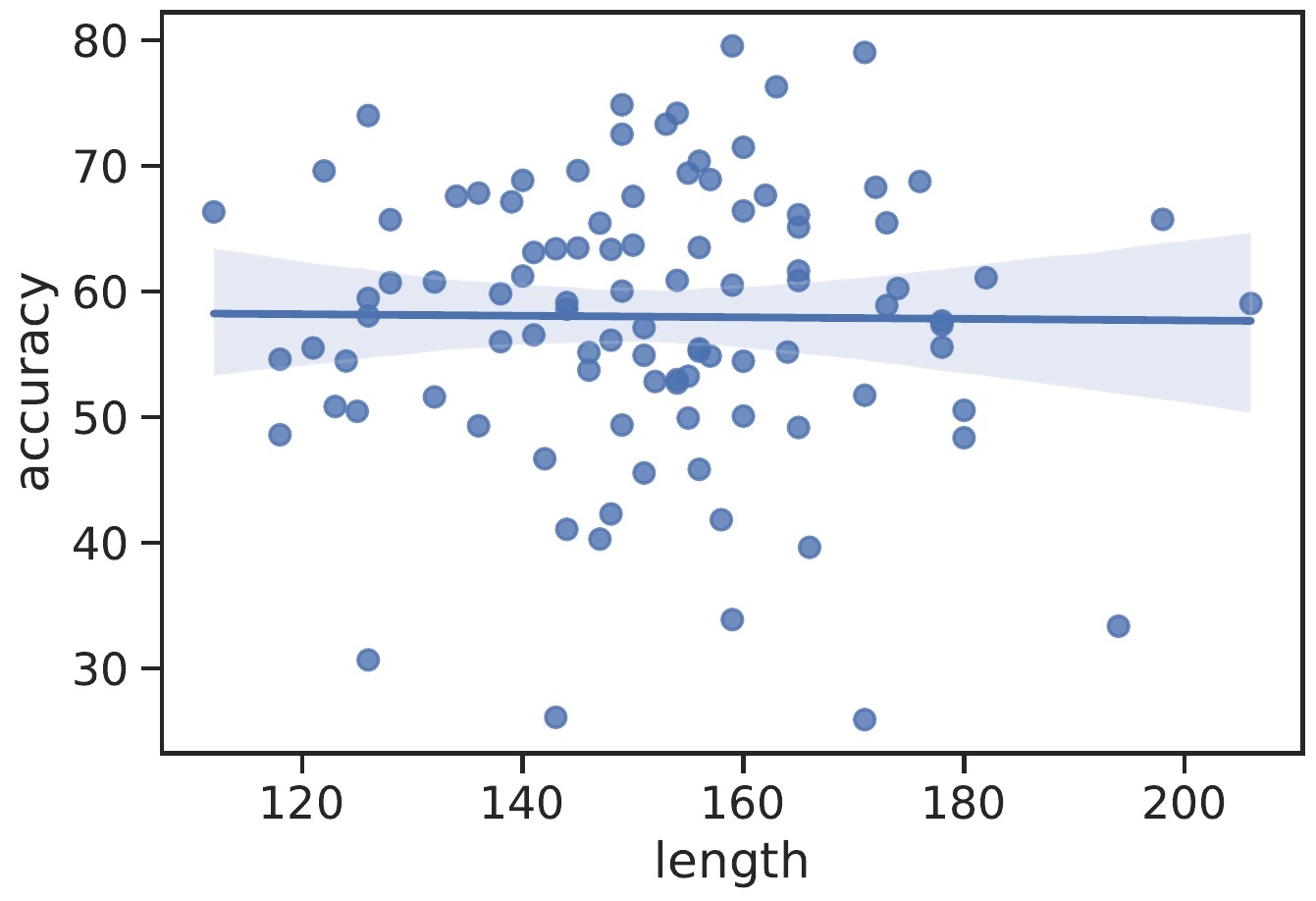}
		\caption{\agnews ($r = -0.01$)}
	\end{subfigure}
	\begin{subfigure}[b]{0.4\textwidth}
		\centering

		\includegraphics[width=\textwidth]{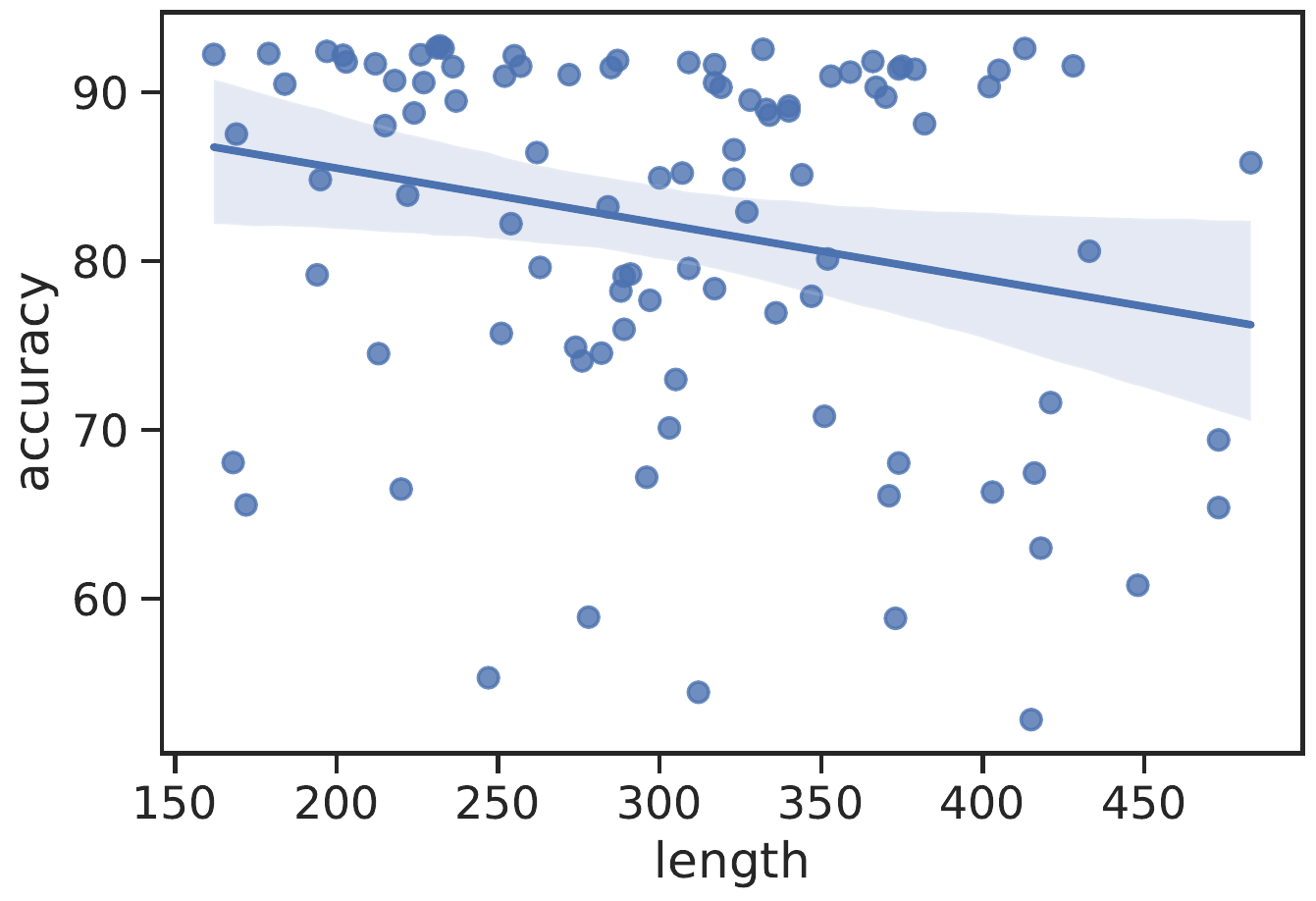}
		\caption{\amazon ($r = -0.23^*$)}
	\end{subfigure}
	\begin{subfigure}[b]{0.4\textwidth}
		\centering

		\includegraphics[width=\textwidth]{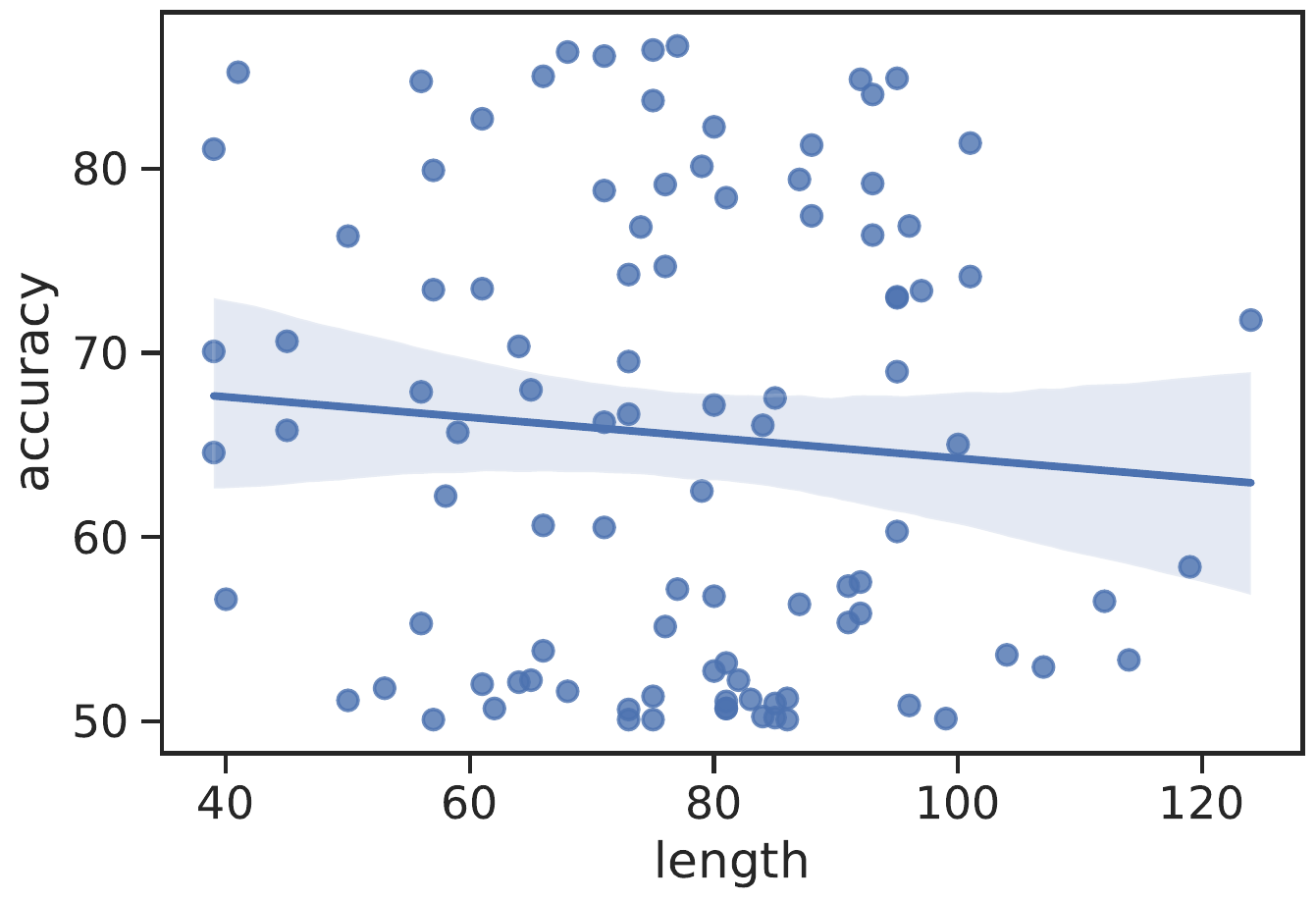}
		\caption{\sst ($r = -0.08$)}
	\end{subfigure}
	\begin{subfigure}[b]{0.4\textwidth}
		\centering

		\includegraphics[width=\textwidth]{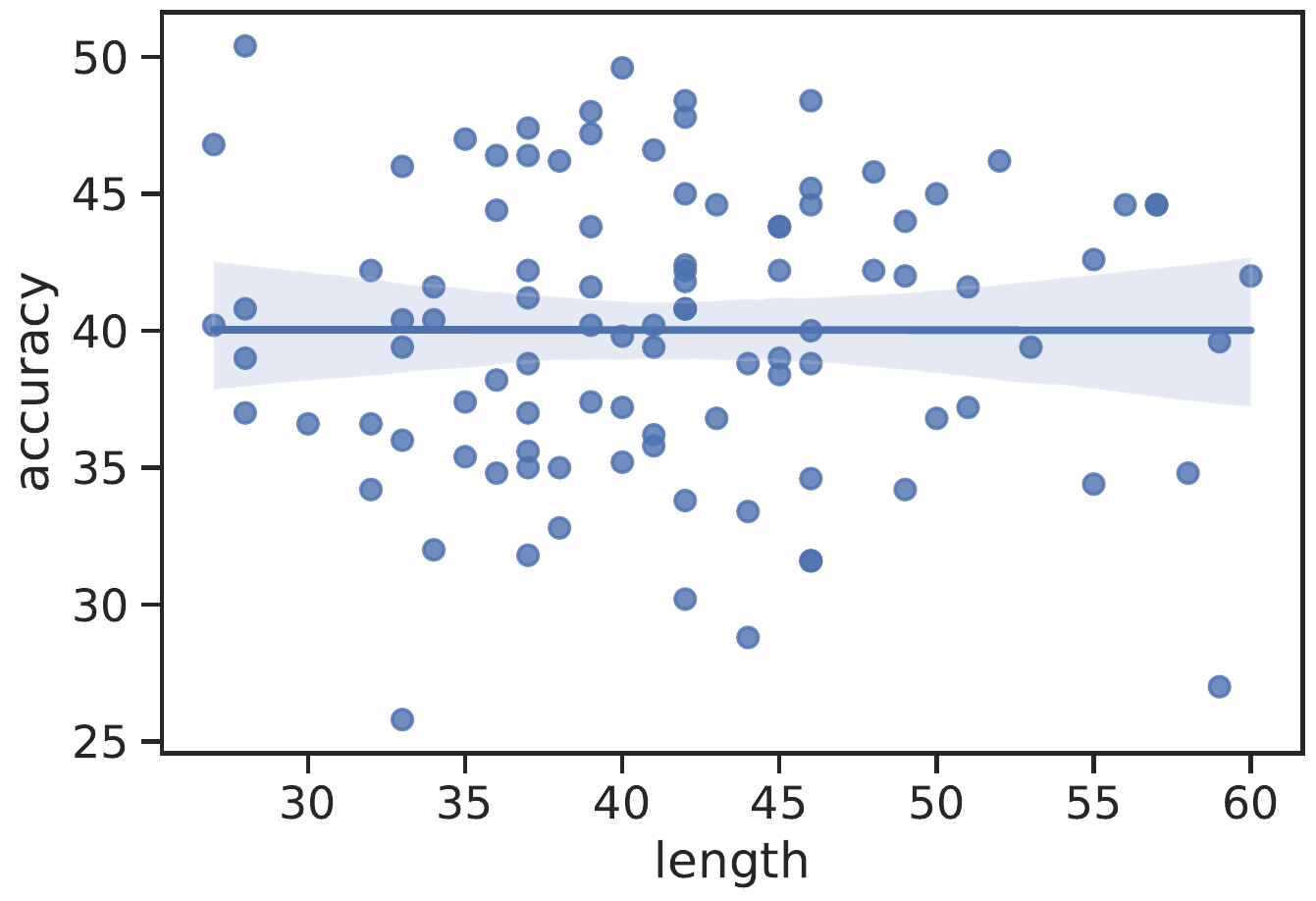}
		\caption{\trec ($r = -0.00$)}
	\end{subfigure}

	\caption{Correlation between length (number of words) of the
		demonstration
		prompt and in-context learning performance across 100 sets of
		randomly
		sample 4-shot demonstration. $*$ indicates a p-value $< 0.05$.}
	\label{fig:length}
\end{figure*}

\end{document}